\documentclass[letterpaper]{article} 
\usepackage[]{aaai2026}  
\usepackage{times}  
\usepackage{helvet}  
\usepackage{courier}  
\usepackage[hyphens]{url}  
\usepackage{graphicx} 
\usepackage{natbib}  
\usepackage{caption} 
\usepackage{algorithm}
\usepackage{algorithmic}
\usepackage{amsmath}
\usepackage{mathtools}
\usepackage{amssymb}
\usepackage{subcaption}
\usepackage{cancel}

\usepackage{newfloat}
\usepackage{listings}
\DeclareCaptionStyle{ruled}{labelfont=normalfont,labelsep=colon,strut=off} 
\floatstyle{ruled}
\newfloat{listing}{tb}{lst}{}
\floatname{listing}{Listing}
\title{More Than Irrational: Modeling Belief-Biased Agents}
\author{
    Yifan Zhu\textsuperscript{\rm 1, \rm 2}, 
    Sammie Katt\textsuperscript{\rm 1, \rm 2}, 
    Samuel Kaski\textsuperscript{\rm 1, \rm 2, \rm 3}
}
\affiliations{
    \textsuperscript{\rm 1}ELLIS Institute Finland\\
    \textsuperscript{\rm 2}Department of Computer Science, Aalto University, Finland\\
    \textsuperscript{\rm 3}Department of Computer Science, University of Manchester, United Kingdom\\
    \{yifan.zhu, sammie.katt, samuel.kaski\}@aalto.fi
    
}

\usepackage{bibentry}

\begin{document}

\maketitle

\begin{abstract}
Despite the explosive growth of AI and the technologies built upon it, predicting and inferring the sub-optimal behavior of users or human collaborators remains a critical challenge.
In many cases, such behaviors are not a result of irrationality, but rather a rational decision made given inherent cognitive bounds and biased beliefs about the world.
In this paper, we formally introduce a class of computational-rational (CR) user models for cognitively-bounded agents acting optimally under biased beliefs. 
The key novelty lies in explicitly modeling how a bounded memory process leads to a dynamically inconsistent and biased belief state and, consequently, sub-optimal sequential decision-making.
We address the challenge of identifying the latent user-specific bound and inferring biased belief states from passive observations on the fly.
We argue that for our formalized CR model family with an explicit and parameterized cognitive process, this challenge is tractable.
To support our claim, we propose an efficient online inference method based on nested particle filtering that simultaneously tracks the user's latent belief state and estimates the unknown cognitive bound from a stream of observed actions.
We validate our approach in a representative navigation task using memory decay as an example of a cognitive bound. 
With simulations, we show that (1) our CR model generates intuitively plausible behaviors corresponding to different levels of memory capacity, and (2) our inference method accurately and efficiently recovers the ground-truth cognitive bounds from limited observations ($\le 100$ steps). 
We further demonstrate how this approach provides a principled foundation for developing adaptive AI assistants, enabling adaptive assistance that accounts for the user's memory limitations.
\end{abstract}

\begin{links}
     \link{Code}{https://github.com/Yifan-Zhu/More-Than-Irrational-Modeling-Belief-Biased-Agents}
\end{links}

\section{Introduction}
In human-AI collaboration, the efficacy of an AI agent depends on its ability to infer the user's goals, beliefs, and future actions from observations of their past behavior. 
Learning such a user model on the fly is particularly challenging, primarily because human behaviors rarely appear to be fully rational. 
Computational rationality (CR) is a universal theoretical framework for understanding such human behavior, positing that humans are rational decision-makers as expected utility maximizers \cite{lewis2014computational, howes2016contextual}, yet our decisions may appear sub-optimal due to latent constraints imposed by subjective utility function, cognitive bounds, and the environment~\cite{oulasvirta_computational_2022, howes_towards_2023, chandramouli2024workflow}. 
Such constraints, especially cognitive bounds, can lead to biased beliefs and sub-optimal behavior.
This happens in everyday life; for example, when going to fetch a misplaced mobile phone, we might check several seemingly irrelevant places simply because our belief about the phone's location is biased due to a faulty or decayed memory. 
An AI assistant capable of inferring the user's memory capacity and tracking their belief state, rather than merely assuming irrationality, would be able to provide more effective assistance.

Computational rationality has shown success in many domains, including eye movements \cite{chen2015emergence, chen2021adaptive}, pointing \cite{ikkala2022breathing}, typing \cite{shi2024crtypist}, and driving \cite{jokinen2021multitasking}.
Surprisingly, little work considers bounded memory resources, and those that do (e.g.~\cite{shi2024crtypist}) are specific to their application.
We argue that a significant amount of irrational behavior can simply be explained by missing or fake memories that lead to incorrect beliefs.
\emph{Biased beliefs, even when acted upon optimally, lead to sub-optimal behavior}.

This work is motivated by a general research question in human-AI collaboration: \emph{how can an AI assistant effectively collaborate with a user whose behaviors seem irrational due to imperfect memory?} 
We investigate how such users with memory-related cognitive bounds can be modeled and learned during online interactions in a partially observable environment. 
To this end, we propose a formal CR framework for explicitly modeling the user's biased belief as the link between latent memory bounds and the user's decisions. 
We posit that these cognitive bounds lead to a subjective belief state that deviates from the objective world.
When the user acts optimally according to this subjective state, they make seemingly sub-optimal decisions. 
This formalism allows the AI to interpret seemingly irrational actions as rational decisions made upon latent bounds and biased beliefs, providing a fundamental user-specific answer to \textit{why} an action was taken and what actions to expect in the future.

Our CR framework is a mathematical solution for inferring the user's latent cognitive bounds and their belief state online; however, it comes with computational challenges. 
We contribute to tackling this issue by first defining a tractable inference setting where the AI has access to the environment model, while the user-specific cognitive parameter remains unknown. 
To solve this joint bound-belief estimation problem, we propose an approximate online Bayesian inference method based on sequential Monte-Carlo techniques. 
We validate our approach in a representative navigation task, the T-maze, using memory decay as an instance of parameterized bounded memory. 
The simulations validate both the expressive power of our CR model in generating intuitive behaviors across various memory capacities and the ability of our inference method to accurately recover the ground-truth bounds from passive observations. 
Building on this validation, we formalize an assistive-POMDP framework in the same task to demonstrate the efficacy of our work in building adaptive AI assistants. 
For example, it can learn to distinguish when a user needs direct action advice due to severe memory decay versus when a subtle memory hint suffices.

\section{Related Work: Computational Rationality} \label{sec:related-work}

Our work can be best understood from the perspective of computational rationality (CR).
This literature views irrationality, such as the behavior of humans, as the result of rational techniques executed under bounded resources.
Seminal papers include~\cite{gershman2015computational,oulasvirta_computational_2022,howes_towards_2023,chandramouli2024workflow}.

A rich body of research in CR has focused on sub-optimal behavior models with latent parameters and their inference.
For example, users in~\citep{kwon2020inverse} maintain an internal model of the environment, though they assume perfect memory given imperfect internal models.
Irrationality has also been explained by bounds on the inference budget~\cite {jacob2023modeling}.
Our work, within this growing field, stands apart as providing a model for bounded memory, which the previous works have not considered much in their general form.

Common in CR, other related work focuses on specific applications, such as gaze prediction~\cite{chen2021adaptive}, pointing~\cite{ikkala2022breathing}, menu search~\cite{chen2015emergence}, and traffic behavior~\cite{jokinen2021multitasking,wang2023modeling}.
To our understanding, work on memory has been limited, though some memory decay was included in user models for using touch screens~\cite{shi2024crtypist,jokinen2021touchscreen}.
As opposed to application-specific solutions, we provide a more general approach that facilitates the modeling of bounded memory in multiple settings.

Of course, for more broadly interested readers, there is an inexhaustive list of works attempting to model decision making and facilitate its inverse problem; yet to our best understanding, they do not support explicit general descriptive models of memory~\cite{jarrett2021inverse,lieder2020resource,alanqary2021modeling}.

\section{Preliminaries: Rational Decision Making}
We model sequential decision-making under uncertainty as a Partially Observable Markov Decision Process (POMDP) \citep{kaelbling1998planning, Sutton2018}. 
Within this framework, we define a fully rational (FR) agent that can maintain an accurate belief state over the latent states via Bayesian filtering and acts optimally to maximize a given objective. 

Formally, a POMDP is defined as a tuple $\mathcal{M} = (\mathcal{S,A},\mathcal{T},\mathcal{R,O}, \Omega, \gamma)$, where $\mathcal{S,A},\Omega$ denote the environment state, action, and observation spaces respectively; $\mathcal{T}(s_{t+1}|s_t, a_t),\mathcal{R}(s_t, a_t),\mathcal{O}(o_{t}|s_{t})$ represent the state transition dynamics, reward function, and observation probability distributions respectively.
At each timestep $t$, the FR agent is in a \emph{latent} world state $s_t \in \mathcal{S}$ and observes $o_{t} \sim \mathcal{O}(o_{t}|s_{t})$, drawn from the observation model $\mathcal{O}: \mathcal{S}  \times \Omega \rightarrow [0,1]$. 
The agent then updates their belief $b_t \in \Delta(\mathcal{S})$ about the environmental states via Bayesian filtering:
\begin{multline}\label{eq-pomdp-belief}
    b_t(s_t) = p(s_t \mid o_t, a_{t-1},b_{t-1}) \\ 
    \propto 
    \mathcal{O}(o_t\mid s_t)\sum_{s_{t-1}}\mathcal{T}(s_t\mid s_{t-1}, a_{t-1})b_{t-1}(s_{t-1}).
\end{multline}
Then, the agent makes an action from its policy $\pi: \Delta(\mathcal{S})\rightarrow \Delta(\mathcal{A})$: $a_t \sim \pi(\cdot\mid b_t)$, and receives reward $r_t \sim \mathcal{R}(s_t, a_t)$; The world then evolves to a new state $s_{t+1}$ stochastically according to the transition dynamics: $s_{t+1} \sim \mathcal{T}(s_{t+1}|s_t, a_t)$. 

\paragraph{Solving a POMDP}
The goal of this FR agent is to learn the optimal policy $\pi_*$ for maximizing the expected return: 
$\pi_*(a\mid b) \propto \exp(\tau Q_*(b,a))$, where $\tau$ is the inverse temperature parameter, and learning $Q_*(b,a)$, the optimal action-value function, is done through RL.
For solution techniques, see~\cite{Sutton2018,shani2013survey,silver2010monte}.

In non-trivial problems, the state space is too large to compute the belief exactly.
The common solution is to approximate the belief with particles.
In \emph{particle filtering}~\cite{gordon1993novel, doucet2001sequential,liu1998sequential}, a distribution $p(x)$ is estimated with $n$ (potentially weighted) Monte-Carlo particles $\{x^i, w^i\}_i^n$ by $p(x) \approx \sum_i w^i \delta_{x^i}(x)$, where $\delta_x(\cdot)$ denotes the Dirac delta mass function at $x$.

\section{Computational Rationality User Modeling}
In this section, we formally define our computational rationality user modeling framework. 
Central to our proposed design is an internal memory process $f_{\theta}$, parameterized by user-specific bounds $\theta \in \Theta$, that explicitly models the agent's memory: an estimate of the observation-action history $h_t \triangleq (\mathbf{o}_{:t}, \mathbf{a}_{:t-1})$.
This function captures, for example, how memories of the past decay over time. 
With this addition, our user model is a POMDP solver using biased beliefs as a result of imperfect memory according to $f_\theta$.

Formally, at time $t$, we define the internal memory over observations and actions \emph{received} up to time $i$ ($t \ge i$) as: 
\begin{equation}
    \tilde{h}_t^{:i} = (\tilde{\mathbf{o}}_{t}^{:i}, \tilde{\mathbf{a}}_{t}^{:i-1}) = (\tilde{o}^1_t,...,\tilde{o}^i_t, \tilde{a}^1_t,...,\tilde{a}^{i-1}_t)
\end{equation}
where each element $\tilde{o}^j_t \in \Omega$ and $\tilde{a}^{j-1}_t \in \mathcal{A}$ is the agent's internal (corrupted) version at time $t$, of the true observation $o_j$ and action $a_{j-1}$ originally received at time $j$. 
The key insight of our CR model is that the memory of past observations can evolve, i.e. $\tilde{o}^j_t \neq \tilde{o}^j_{t-1}$. 
For notational simplicity, we denote the complete internal memory $\tilde{h}^{:t}_t$ at time $t$ as $\tilde{h}_t$. 

The agent's internal state $\tilde{h}_t$ evolves according to the internal stochastic dynamics function $f_{
\theta}: \Omega^{t-1} \times \mathcal{A}^{t-2} \times \Omega \times \mathcal{A} \rightarrow \Delta(\Omega^{t} \times \mathcal{A}^{t-1})$ that maps the previous internal history $\tilde{h}_{t-1}$ and new observation pair ($o_t, a_{t-1}$) to a distribution over the current cognitive state: 
\begin{equation} \label{eq-f}
    \tilde{h}_{t} \sim f_{\theta}(\tilde{h}_{t-1}, o_{t}, a_{t-1}), 
\end{equation}
which is parameterized by the cognitive bound $\theta$. 

For intuitions, recall the phone-searching example. 
A limited memory capacity can result in forgetfulness, resulting in corrupted memory and, consequently, biased beliefs.
For example, humans frequently forget about their last interaction(s) with their phone and, hence, have incorrect beliefs about its location.
The function $f_{\theta}$ provides the formal mechanism for this corruption process. 

Unlike rational agents, a CR agent computes its belief state $\tilde{b}_t \in \Delta(\mathcal{S})$ via Bayes' rule by conditioning on the corrupted internal memory of the history:
\begin{equation} \label{eq-biased-belief}
    \begin{aligned}
        \tilde{b}_t  &\triangleq p(s_t \mid \tilde{\mathbf{o}}_{t}^{:t}, \tilde{\mathbf{a}}_t^{:t-1})
                        \propto \sum_{\textbf{s}_{:t-1}} p(\textbf{s}_{:t}, \tilde{\mathbf{o}}_{t}^{:t} \mid \tilde{\mathbf{a}}_t^{:t-1}) \\
                        & = \sum_{\textbf{s}_{:t-1}} p(s_0) \mathcal{O}(\tilde{o}^0_{t} \mid s_0) \\& \qquad \qquad \times\prod_{i=1}^t \mathcal{O}(\tilde{o}_{t}^i \mid s_i) \mathcal{T}(s_i \mid s_{i-1}, \tilde{a}_{t}^{i-1}).
    \end{aligned}
\end{equation}
Because the Markov property no longer holds for the biased belief, computing it involves marginalizing out the agent's entire, potentially flawed, memory sequence, rather than simply a one-step update based on the previous belief in Equation \eqref{eq-pomdp-belief}.
Namely, when memory modifies $\tilde{o}_t^i$, the belief state shifts retroactively.
Equation \eqref{eq-biased-belief} along with Equation \eqref{eq-f} shows that the cognitive bound $\theta$ introduces bias into the belief by systematically corrupting the elements in $\tilde{h}_t$ through $f_{\theta}$. 
This non-trivial formalism allows for explicitly capturing human-like decision-making that often involves ``replaying'' or ``re-evaluating'' memories. 

The CR agent is internally rational, acting optimally based on their subjective beliefs:
\begin{equation} \label{eq:biased-policy}
    \pi_*(a \mid \tilde b;\theta)
= \frac{\exp(\tau\, Q_*(\tilde{b}, a;\theta))}
       {\sum_{a'\in\mathcal{A}} \exp(\tau\, Q_*(\tilde{b}, a';\theta))},
\end{equation}
where $\tau$ is the inverse temperature parameter, $Q_*(\cdot;\theta)$ is the optimal action-value function learned based on the agent's biased beliefs.
The CR agent's action $a_{\text{CR}} \sim \pi_*(a\mid \tilde{b};\theta)$ is rational from its own perspective, yet for external observers with access to the objective belief $b^*$, the exact action may appear sub-optimal (i.e. $Q_*(b^*, a_{\text{CR}}) < Q_*(b^*, a_{\text{FR}}))$. 
This framework thus provides a principled and elegant way to model sub-optimal behavior stemming from biased beliefs as a result of latent cognitive bounds.

\section{Online Inference of Bounds and Beliefs}
A direct result of the CR user model above is the ability to simulate agents with bounded and evolving memory.
Most useful applications necessitate solving the inverse problem: inferring the latent bound from observed behavior.
In this section, we present and address the technical challenge of learning such user models by inferring $\theta$ online. 
We first present a well-defined problem setting, then introduce an online inference approach based on nested particle filtering. 

\subsection{Problem Setting}
We are interested in estimating an agents' bound parameter $\theta$ online given a passively observed action-observation history ($h_t \triangleq (\mathbf{o}_{:t}, \mathbf{a}_{:t-1})$).
Given that we have a likelihood model --- the user model described above (Equation ~\eqref{eq:biased-policy}) --- it is natural to assume a uniform prior over the parameter of interest $p(\theta)$ and consider the Bayesian task of inferring its posterior.
It is natural to consider the joint distribution over parameter and internal state (here $\tilde{h}_{t-1}$), as the following derivation shows:
\begin{equation} \label{eq-ai-belief}
    \begin{aligned}
    &p(\tilde{h}_{t-1}, \theta \mid h_t) \\
    & = \mathrlap{p(\tilde{h}_{t-1}, \theta \mid h_{t-1}, a_{t-1}, o_t)} \\
    & \propto \sum_{\tilde{h}_{t-2}} p(\tilde{h}_{t-1}, \tilde{h}_{t-2}, a_{t-1}, \theta \mid h_{t-1}, o_t) \\
    & = p(a_{t-1} \mid \tilde{h}_{t-1}) \\ 
    & \qquad \times \sum_{\tilde{h}_{t-2}} p(\tilde{h}_{t-1} \mid \tilde{h}_{t-2}, h_{t-1}, \theta )p(\tilde{h}_{t-2}, \theta \mid h_{t-1})  \\ 
    & = \overbrace{p(a_{t-1} \mid \tilde{h}_{t-1})}^{\text{user action likelihood}} \quad \sum_{\tilde{h}_{t-2}} \bigg [ \overbrace{p(\tilde{h}_{t-2}, \theta \mid h_{t-1})}^{\text{recursive}} \\
    & \qquad \times f_{\theta}(\tilde{h}_{t-1} \mid \tilde{h}_{t-2}, o_{t-1}, a_{t-2}) \bigg ]. 
    \end{aligned}
\end{equation}
Let us consider these three terms.
First, the memory model $f_{\theta}$ is assumed to be given (for example memory decay, as used in our experiments).
The ``recursive'' term is the belief quantity computed at the previous timestep and thus given.
The first term, the ``user action likelihood'', reflects the core idea of this work and represents the probability that a CR agent takes action $a$ given a (biased) belief resulting from corrupted memory $\tilde{h}_{t-1}$.
In practice, this is computed by deriving the biased belief $\tilde{b}$ given the corrupted memory and computing the optimal action given this belief (recall Equations~\eqref{eq-biased-belief} \&~\eqref{eq:biased-policy}), i.e. $p(a_{t-1}\mid \tilde{h}_{t-1}) = \pi_*(a_{t-1}\mid \tilde{b}_{t-1};\theta)$ with $\tilde{b}_{t-1} = p(s \mid \tilde{h}_{t-1})$.
To compute this, we assume knowledge of the system dynamics ($\mathcal{T}$ \& $\mathcal{O}$) and the user's bounded memory model family $f_{\theta}$ (though not their latent parameter $\theta$!).
In practice, we precompute these policies.

This derivation formally exposes the non-trivial computational structure of the inference problem.
The computational intractability of the belief update is clear: the summation, $\sum_{\tilde{h}_{t-2}}$, is over the high-dimensional space of (corrupted) memory sequences and grows in time.
This, we address next.

\subsection{Online Inference via Nested Particle Filtering}
In light of the computation challenges and the problem structure of estimating a static parameter $\theta$ alongside a dynamic latent state $\tilde{h}_{t-1}$, we propose to approximate the posterior over $(\tilde{h}_{t-1}, \theta)$ with \emph{nested} particle filtering (NPF). 
We find this technique particularly effective when a joint distribution is best represented as a marginal and conditional distribution: $p(\tilde{h}_{t-1}, \theta) = p(\theta) p(\tilde{h}_{t-1} | \theta)$.
Additionally, since in our setting the parameter space $\Theta$ is small and the policy likelihood $\pi_*(\cdot;\theta)$ is expensive to learn and thus realistically must be precomputed, NPF suits naturally here over other inference methods, such as Particle MCMC \cite{andrieu2010particle}, where the latter samples novel $\theta$ and requires computing policies online.

In practice, NPF maintains $N_{\theta}$ (potentially weighted) particles $\{\theta^i\}_{i=1}^{N_{\theta}}$ and, \emph{for each particle} $\theta^i$, $N_{\tilde{h}}$ conditional particles $\{\tilde{h}_{t-1}^{(i,j)} \}_{j=1}^{N_{\tilde{h}}}$.
The posterior, following standard Monte-Carlo formalisms, is then approximated as follows:
\begin{align}
    p(\theta \mid h_t) & \approx \sum_{i=1}^{N_{\theta}}w^i \delta_{\theta^i}(\theta) \\
    p(\tilde{h}_{t-1} \mid h_t, \theta^i) & \approx \sum_{j=1}^{N_{\tilde{h}}}w^{(i,j)} \delta_{\tilde{h}_{t-1}^{(i,j)}}(\tilde{h}_{t-1})\label{eq-npf-phi}
\end{align}
We utilize particle filtering to update the posterior given a new action-observation pair.
The key step in this process is the weight update: the weight of each outer particle, $w^i$, is updated according to the likelihood of explaining the user's most recent action $a_{t-1}$, estimated by the inner particle filter associated with $\theta^i$.
Our contribution, regarding inference methodology, is the specific formulation of this likelihood computation within our CR framework. 
As detailed in Algorithm \ref{alg:alg-npf}, this likelihood computation requires simulating the user's internal decision-making process for each particle $(i,j)$: first, each internal state particle $\tilde{h}^{(i,j)}$ is updated using our cognitive process $f_{\theta^i}$, and the corresponding biased belief $\tilde{b}^{(i,j)}$ is computed (line 4 and 5).
Then the weights of all the particles are updated according to their likelihood of producing the observed action $\pi_*(a \mid \tilde{b}^{(i,j)};\theta^i)$ (line 6).
Lastly, the weights are re-normalized (line 10 to 12).

With NPF estimation, it is possible to infer a user's bounds and track their biased belief online: the computational cost of exact inference in Equation \eqref{eq-ai-belief} is $O(|\mathcal{S}|^t t!)$, while our inference method only costs $O(N_{\theta}N_{\tilde{h}}t|\mathcal{S}|)$ (biased belief computation in Equation \eqref{eq-biased-belief} costs $O(t|\mathcal{S}|)$), which allows for real-time inference.

\begin{algorithm}[tb]
\caption{Nested Particle Filter Update $p(\tilde{h}_{t-1}, \theta \mid h_t)$}\label{alg:alg-npf}

\textbf{Input}: Previous Particles: $\big \{\theta^i, w^i, \{ \tilde{h}_{t-2}^{(i,j)}, w^{(i,j)} \}_{j=1}^{N_{\tilde{h}}} \big \}_{i=1}^{N_\theta}$  \\
\hspace*{2.8em} History $h_t$: $(\mathbf{o}_{:t}, \mathbf{a}_{:t-1})$
\begin{algorithmic}[1]
    \STATE // update particles:
    \FOR{$i = 1,\dots,N_\theta$}
        \FOR{$j = 1,\dots,N_{\tilde{h}}$}
            \STATE Sample $\tilde{h}_{t-1}^{(i,j)} \sim f_{\theta^i}(\tilde{h}_{t-2}^{(i,j)}, a_{t-2}, o_{t-1})$
            \STATE Compute $\tilde{b}_{t-1}^{(i,j)} \leftarrow p(s \mid \tilde{h}_{t-1}^{(i,j)})$
            \STATE Evalute likelihood $L^{(i,j)} \leftarrow \pi_*(a_{t-1} \mid \tilde{b}_{t-1}^{(i,j)}; \theta^i)$
        \ENDFOR
    \ENDFOR
    \STATE // update and normalize weights:
    \STATE $w^{(i,j)} \leftarrow w^{(i,j)} L^{(i,j)}$
    \STATE $w^i \leftarrow \frac{ w^i \sum_{j=1}^{N_{\tilde{h}}}w^{(i,j)} }{ \sum_{i'} w^{i'} \sum_{j=1}^{N_{\tilde{h}}}w^{(i',j)}}$ 
    \STATE $w^{(i,j)} \leftarrow \frac{w^{(i,j)}}{\sum_{j'} w^{(i,j')}}$ 
    \STATE \textbf{return} updated particles $ \big \{\theta^i, w^i, \{ \tilde{h}_{t-1}^{(i,j)}, w^{(i,j)} \}_j^{N_{\tilde{h}}} \big \}_{i=1}^{N_\theta} $
\end{algorithmic}
\end{algorithm}

\begin{figure}[t]
\centering
\includegraphics[width=0.4\columnwidth]{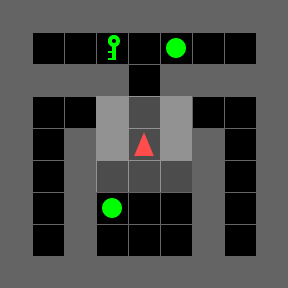}
\caption{Rendering of the T-maze simulation task. The agent (red triangle) must explore the bottom room to find the target object (ball in this case), memorize it, and navigate to the corresponding terminal state (one of the arms of the ``T'', here the right-side) to succeed.}
\label{fig:tmaze}
\end{figure}

\begin{figure*}[ht]
    \centering
    \begin{subfigure}[b]{0.24\linewidth}
        \centering
        \includegraphics[width=\linewidth]{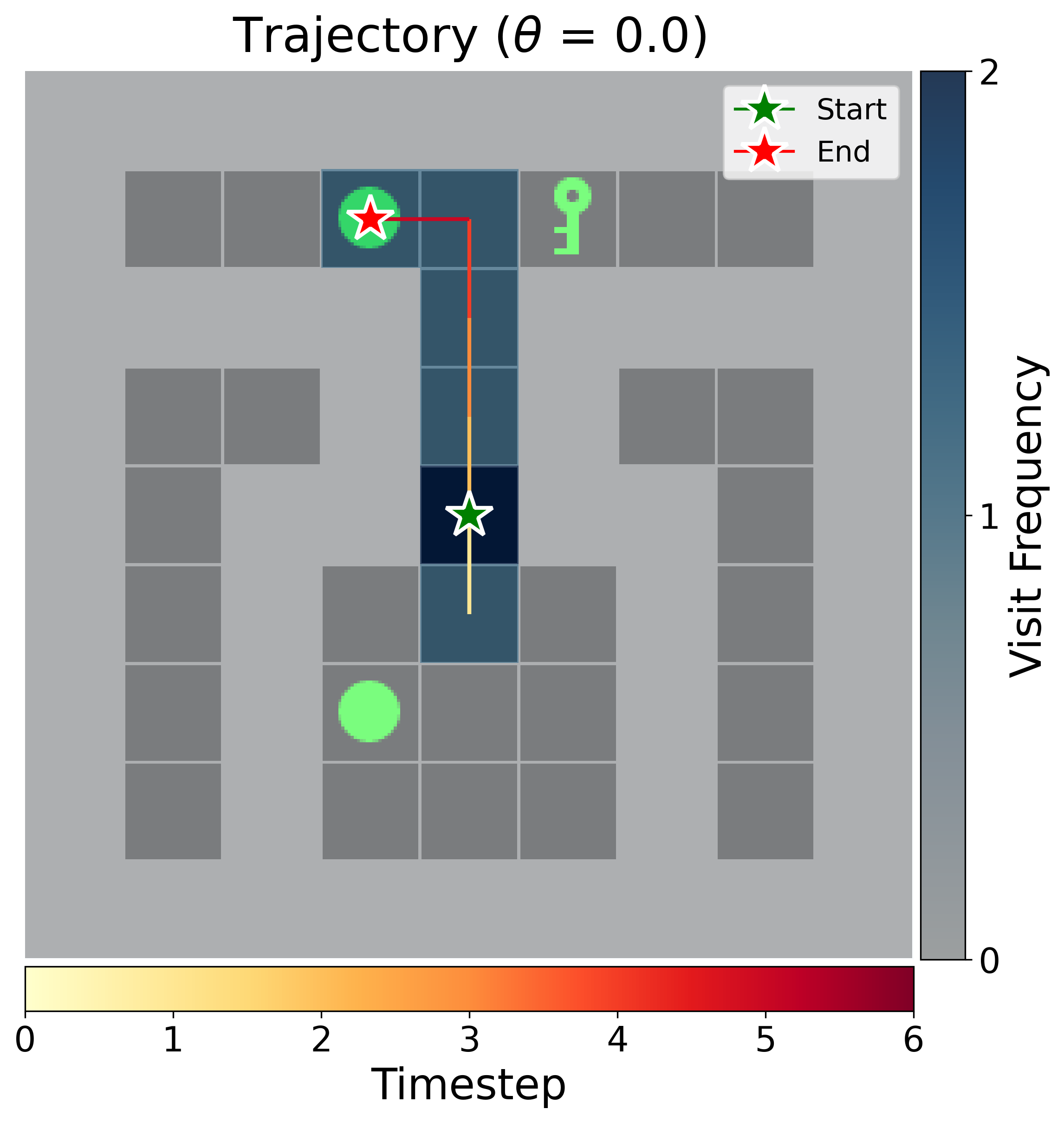}
        \caption{Perfect Memory ($\theta=0.0$)}
        \label{fig:traj_perfectmem}
    \end{subfigure}
    \hfill 
    \begin{subfigure}[b]{0.24\linewidth}
    \centering
        \includegraphics[width=\linewidth]{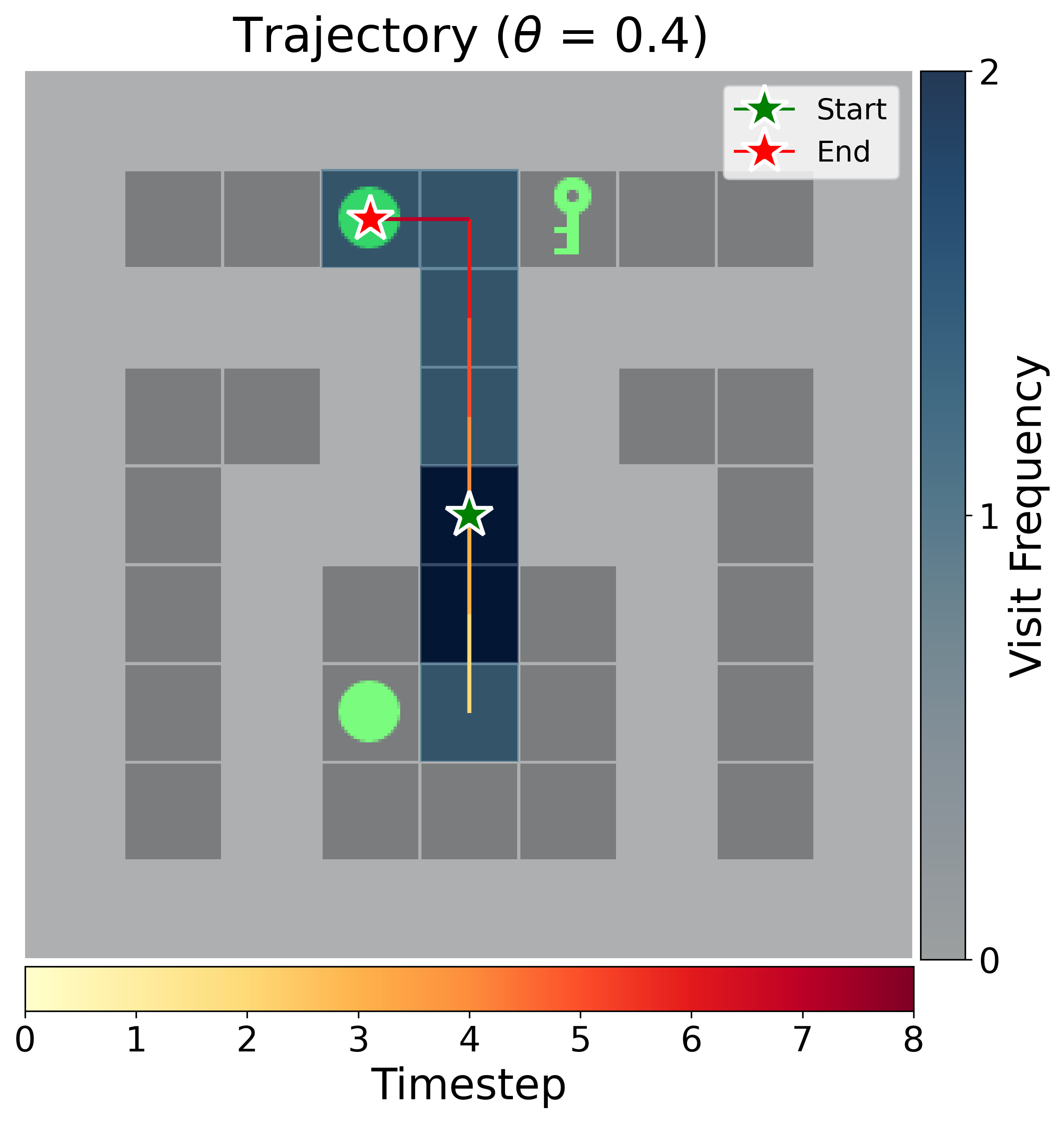}
        \caption{Imperfect Memory ($\theta=0.4$)}
        \label{fig:traj_imperfect_0.4}
    \end{subfigure}
    \hfill 
    \begin{subfigure}[b]{0.24\linewidth}
    \centering
        \includegraphics[width=\linewidth]{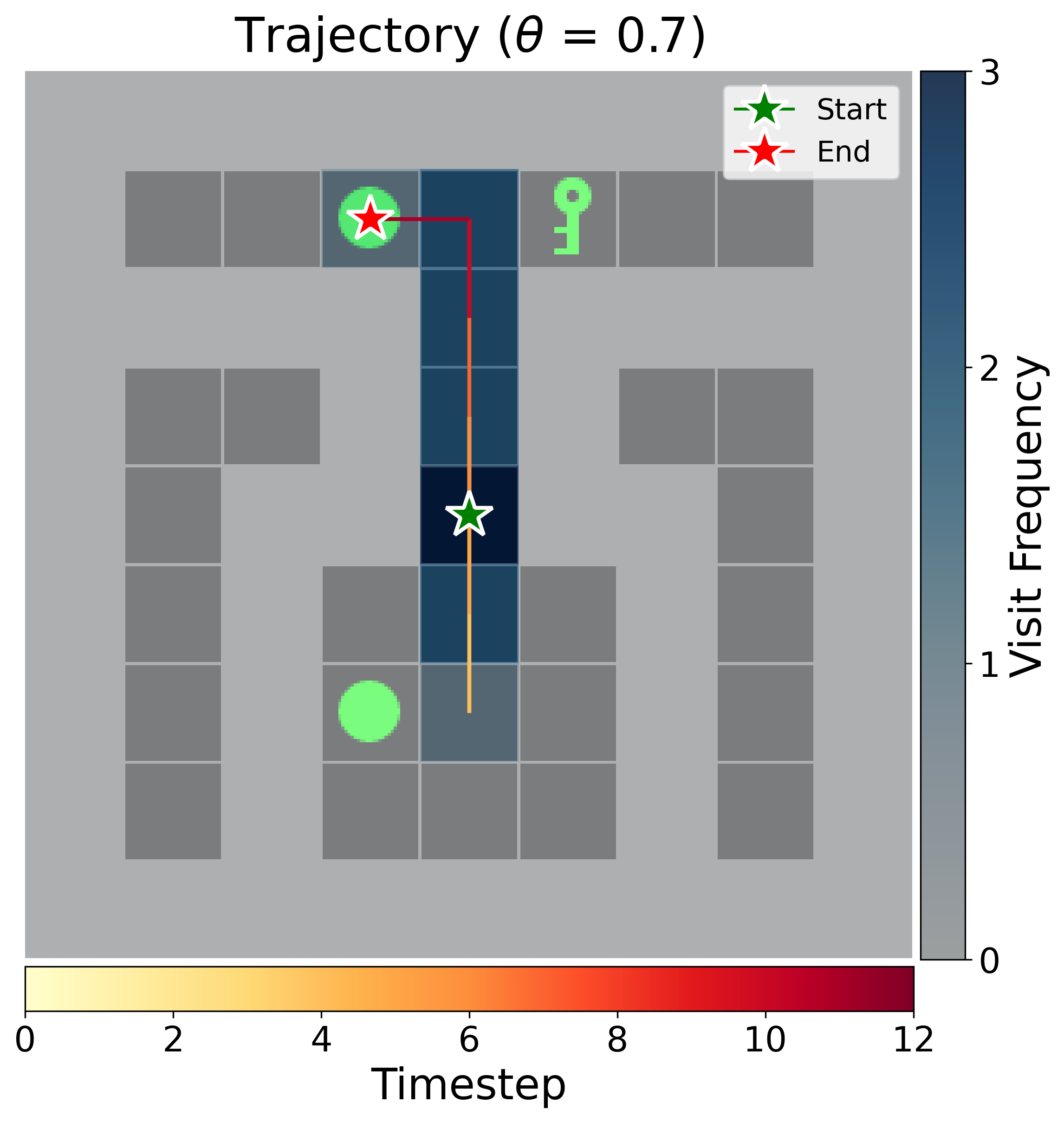}
        \caption{Imperfect Memory ($\theta=0.7$)}
        \label{fig:traj_imperfect_0.7}
    \end{subfigure}
    \hfill 
    \hfill 
    \begin{subfigure}[b]{0.24\linewidth}
    \centering
        \includegraphics[width=\linewidth]{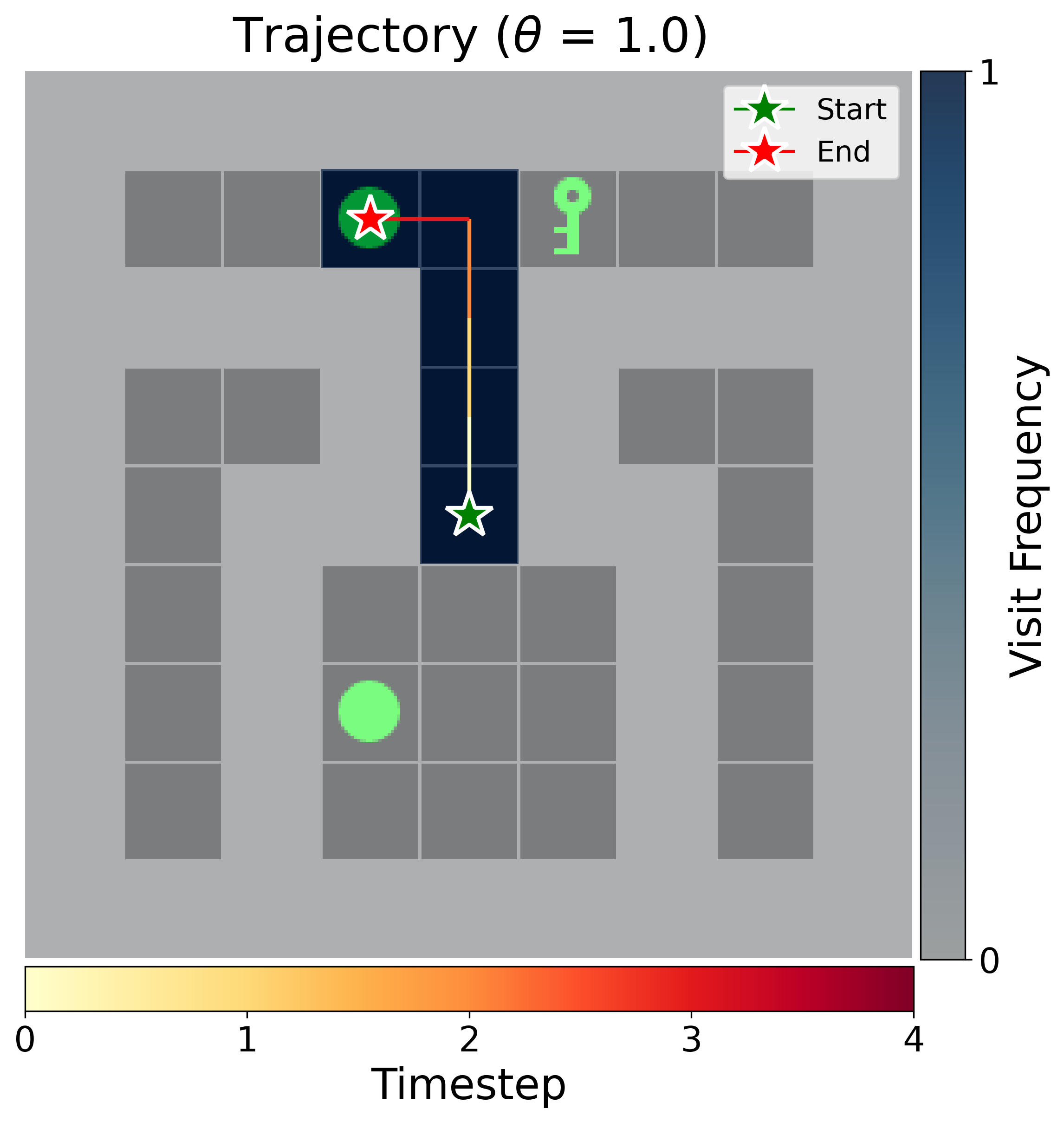}
        \caption{No Memory ($\theta=1.0$)}
        \label{fig:traj_nomem}
    \end{subfigure}
    \caption{Representative trajectories generated by our CR model for agents with different memory bound parameters $\theta$. 
    The color of the trajectory indicates the temporal order of the steps (brighter is earlier), and the colored grids indicate the visit frequency.
    The agent starts in the hallway (green star) and reaches one of the terminal states (red star). (a) With perfect memory, the agent acts optimally by going down to find the object and directly proceeds to the correct terminal state. (b) With a 40\% chance of losing memory, the agent learned to collect more observations on the object. (c) With a 70\% chance of losing memory, the agent learned to recheck the found position after it appeared to have forgotten. (d) With no memory, the agent learned to take a random guess without exploring the environment. }
    \label{fig:exp_crtraj}
\end{figure*}

\section{Experiments}
In this section, we present a series of simulation-based experiments designed to provide a foundational proof-of-concept validation for our core claims.
Specifically, we aim to answer the following key questions through simulations:
\begin{enumerate}
    \item Can our proposed CR user model generate a spectrum of intuitively plausible, sub-optimal behaviors that correspond to different levels of cognitive bounds?
    \item Can our online inference method accurately and efficiently infer the user's latent cognitive bounds and state from only passive observations?
    \item Does the inferred cognitive bound serve as useful information for a downstream adaptive assistance task?
\end{enumerate}
To this end, we take a typical navigation task that requires memory use as a simple clear testbed for our validation.

\subsection{Simulation Task: T-maze}
All our experiments are performed on a simulated grid-world partial observable navigation task based on the T-maze, a classical test for working memory.
As illustrated in Figure \ref{fig:tmaze}, the agent (red triangle) starts in a hallway and is tasked with navigating to the goal, one of the two terminating locations at the top. 
The optimal behavior first explores the bottom room, observes and memorizes the object (here a ball), and then proceeds through the hallway to reach the location with the same object as seen.
The hidden state contains both the agent's current position and the target object; The agent can choose to go to any of the four directions, or simply stay in place; The observation is the 3 by 3 grid around the agent; The transition function is deterministic; reaching the correct terminal state generates a reward of $1-0.9\times \frac{\text{timesteps}}{\text{maxsteps}}$, in all other scenarios the reward is 0.
This task is representative in our case as it creates a clear credit assignment problem that can only be solved by maintaining a specific piece of information in memory over a period of time, and we expect varying behaviors depending on the quality of the memory model.

\subsection{Validation of the CR User Model}
\subsubsection{Setup}
We start by studying the expressive power of our CR user model.
We instantiate a CR agent with a cognitive process $f_\theta$ designed to capture the gradual decay of memory.
We implement this as having a $p=\theta$ chance of corrupting the observation of the target object.
Specifically, at each step $t$, after the new observation $(o_t,a_{t-1})$ is added to the buffer, for each observation of the target object in $\tilde{h}_t$, there is a probability of $\theta$ that the seen object is replaced by a default value.
The parameter $\theta$ here thus represents the memory decay rate, where an agent with $\theta=0.0$ has perfect memory and the one with $\theta = 1.0$ forgets the object seen immediately.
To evaluate the behavior trajectories generated by our CR user model in this case, we pre-trained optimal policies $\pi_*(a\mid \tilde{b};\theta)$ for a range of $\theta$ values using Proximal Policy Optimization (PPO) \cite{schulman2017proximal} and visualized the corresponding trajectories.
The goal is to verify that our model can generate a range of distinct and cognitively plausible behaviors by varying the memory decay rate $\theta$.

\subsubsection{Results}
We present 4 representative trajectories in Figure \ref{fig:exp_crtraj}, illustrating the behavioral pattern of CR agents having perfect memory, imperfect memory, and no memory: 
\begin{itemize}
    \item Perfect Memory ($\theta = 0.0$): with no decay, the agent learned to act optimally by taking the shortest path to complete the task successfully: one step down to observe the object, then proceed to the correct terminal state.
    \item Imperfect Memory ($\theta \in \{0.4, 0.7\}$): agents with various levels of memory decay exhibit classic sub-optimal but highly intuitive behavioral patterns. With a moderate decay rate $\theta=0.4$, the agent learned to collect enough observations for robust memorization. With worse memory capability ($\theta = 0.7$), the agent shows a ``forget-recheck'' pattern: they forgot the seen object before making the turning decision, and went down to re-check the object.
    \item No Memory ($\theta = 1.0$): with no memory, the agent is unable to solve the task strategically. It does not even waste time exploring; instead, it makes a random turn.
\end{itemize}
This result shows that our CR model is qualitatively capable of generating a series of diverse and intuitively plausible behavioral patterns given different memory-bound values.
The simple yet interpretable mechanism of memory decay is sufficient to generate a spectrum of behaviors that are not merely random and irrational, but are actually sub-optimal in a structured and plausible way. 

\begin{figure}[h]
\centering
\includegraphics[width=\columnwidth]{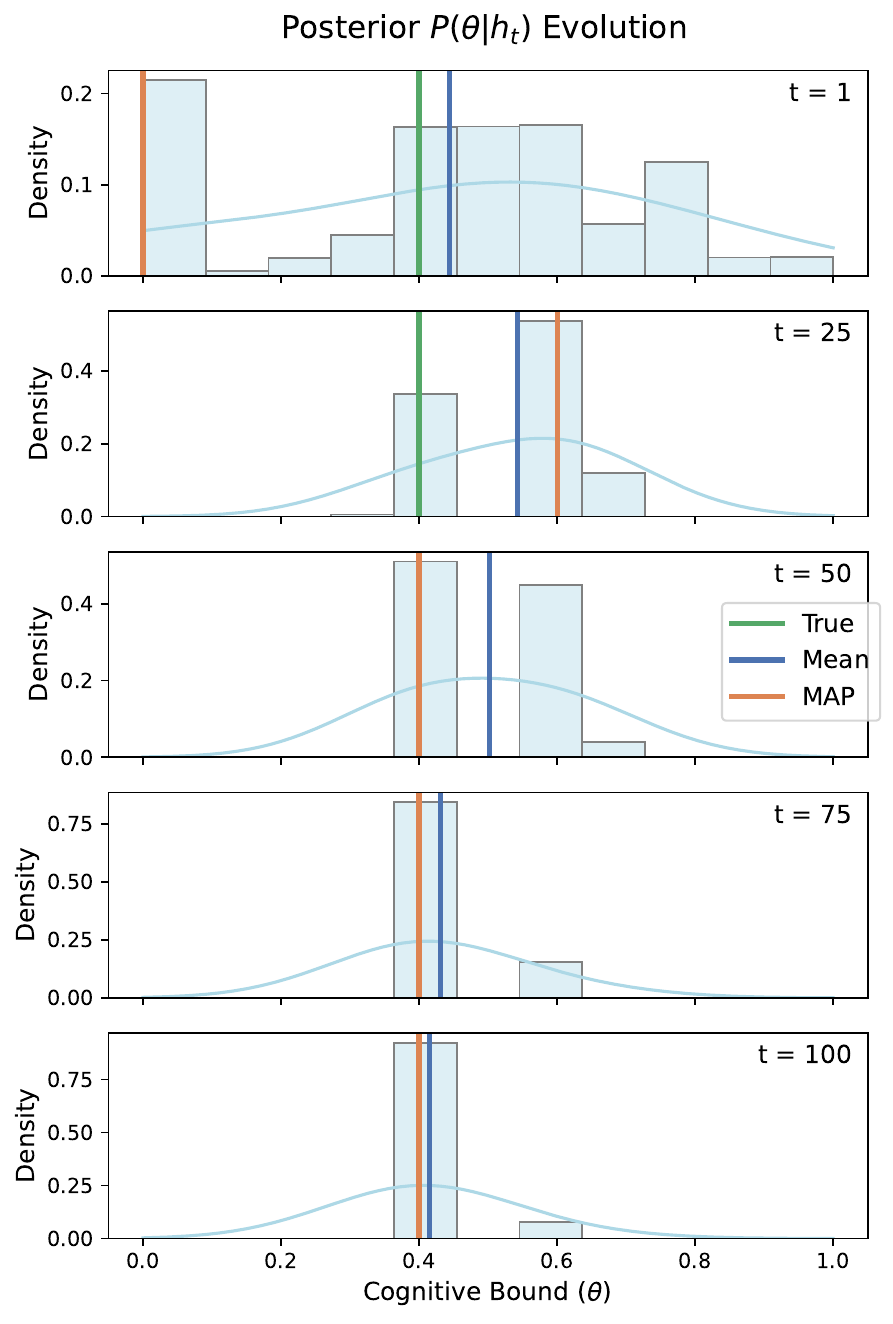}
\caption{Visualization of $p(\theta|h_t)$ evolving over time where $\theta_{\text{true}}=0.4$ and $\tau=3.0$. Each panel shows the estimated posterior at representative timesteps, where the histogram represents the distribution of the weighted particles and the curve is the kernel density estimate used to approximate the posterior. The posterior mean (blue line) and MAP estimate (red line) rapidly converge towards the true value (green line) as more observations are made.}
\label{fig:posterior_evolution}
\end{figure}
\subsection{Validation of Online Bound-Belief Inference}
\subsubsection{Setup}
To evaluate our online inference method, we run Algorithm~\ref{alg:alg-npf} on action-observation trajectories of a CR agent in $100$ timesteps across multiple episodes.
For each simulation, we sample a ground-true $\theta_{\text{true}}$ uniformly from $\{0.0, 0.1,...,1.0\}$ for the CR agent. 
Then, we assume same uniform prior over $\theta_{\text{true}}$
and use our method to infer the joint posterior $p(\tilde{h}_{t-1}, \theta \mid h_t)$ with $N_{\theta}$ and $N_{\tilde{h}}$ particles.

We evaluate the performance of our inference method with two metrics: the Posterior Mean (PM) error ($|\mathbb{E}[\theta \mid h_t] - \theta_{\text{true}}|$), which is the absolute difference between the mean of the estimated posterior $p(\theta \mid h_t)$ and $\theta_{\text{true}}$, and the Maximum a Posteriori (MAP) error ($|\arg\max_{\theta}p(\theta\mid h_t) -\theta_{\text{true}}|$), which is the absolute difference between the most probable estimated value and $\theta_{\text{true}}$. 
All results are reported as means and standard errors over runs with all $\theta_{\text{true}}$ and 5 different random seeds, and sensitivity analysis of hyperparameter $N_{\tilde{h}}$ and $\tau$ is available in the extended version, which can be found in our code repository.

\begin{figure}[ht]
\centering
\includegraphics[width=\columnwidth]{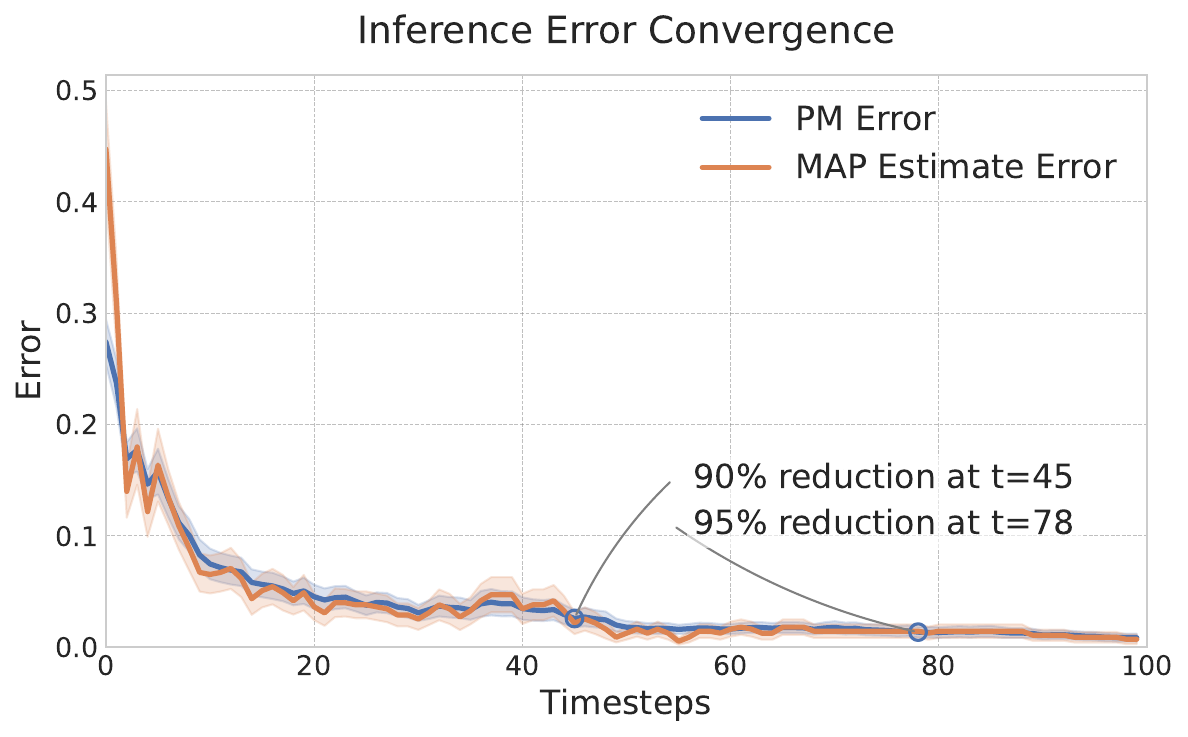}
\caption{Inference error convergence: the PM error and MAP error over time (mean $\pm$ standard error over 5 seeds and all $\theta_{\text{true}}$). Relative to the initial error at $t=1$, the PM error decreased by 90\% and 95\% at $t=45$ and $t=78$ respectively, demonstrating the inference efficiency. }
\label{fig:convergence}
\end{figure}
\subsubsection{Results}
We present a qualitative plot of the inference updates and a quantitative convergence plot to demonstrate that our method is both accurate and efficient for identifying the user's latent parameter $\theta$ from passive observations.

Figure \ref{fig:posterior_evolution} visualizes the evolution of the posterior distribution $p(\theta \mid h_t)$ at representative timesteps (steps 1, 25, 50, 75, and 100) where $\theta_{\text{true}}=0.4$. 
After one-step inference from a uniform prior, the estimated posterior shifts dramatically, indicating that the user's initial action is highly informative, which may depend on the task design.
With more observations, the estimation concentrates around plausible hypotheses, i.e. $\{0.4,0.6\}$, showing the challenge of identifiability: different cognitive bounds can lead to similar actions.
As evidence accumulates, the estimation converges gradually, and our method identifies the true bound from competing hypotheses.
The figure shows how our method reduces uncertainty in identifying $\theta$ over time with more observations of the environment and the user's actions.

To quantify the accuracy and efficiency of the process, we aggregated the results across all ground-truth $\theta$ conditions and 5 random seeds where $\tau=3.0$. 
Figure \ref{fig:convergence} plots the evolution of the PM error and MAP error over 100 steps, confirming two key properties of our method:
\begin{enumerate}
    \item Accuracy: Both error metrics rapidly converge to near-zero, with the final PM error being 0.0087 $\pm$ 0.0035 (standard error), demonstrating method effectiveness.
    \item Data Efficiency: The majority of the error reduction occurs within the first 20-30 steps ($\sim$ 2-3 episodes) of observations, and the PM error is reduced by $90\%$ after 45 steps and by $95\%$ after 78 steps (baseline being PM error at $t=1$), demonstrating the applicability potential of our approach in real-time assistance.
\end{enumerate}

\begin{figure}[hpt]
\centering
\includegraphics[width=\columnwidth]{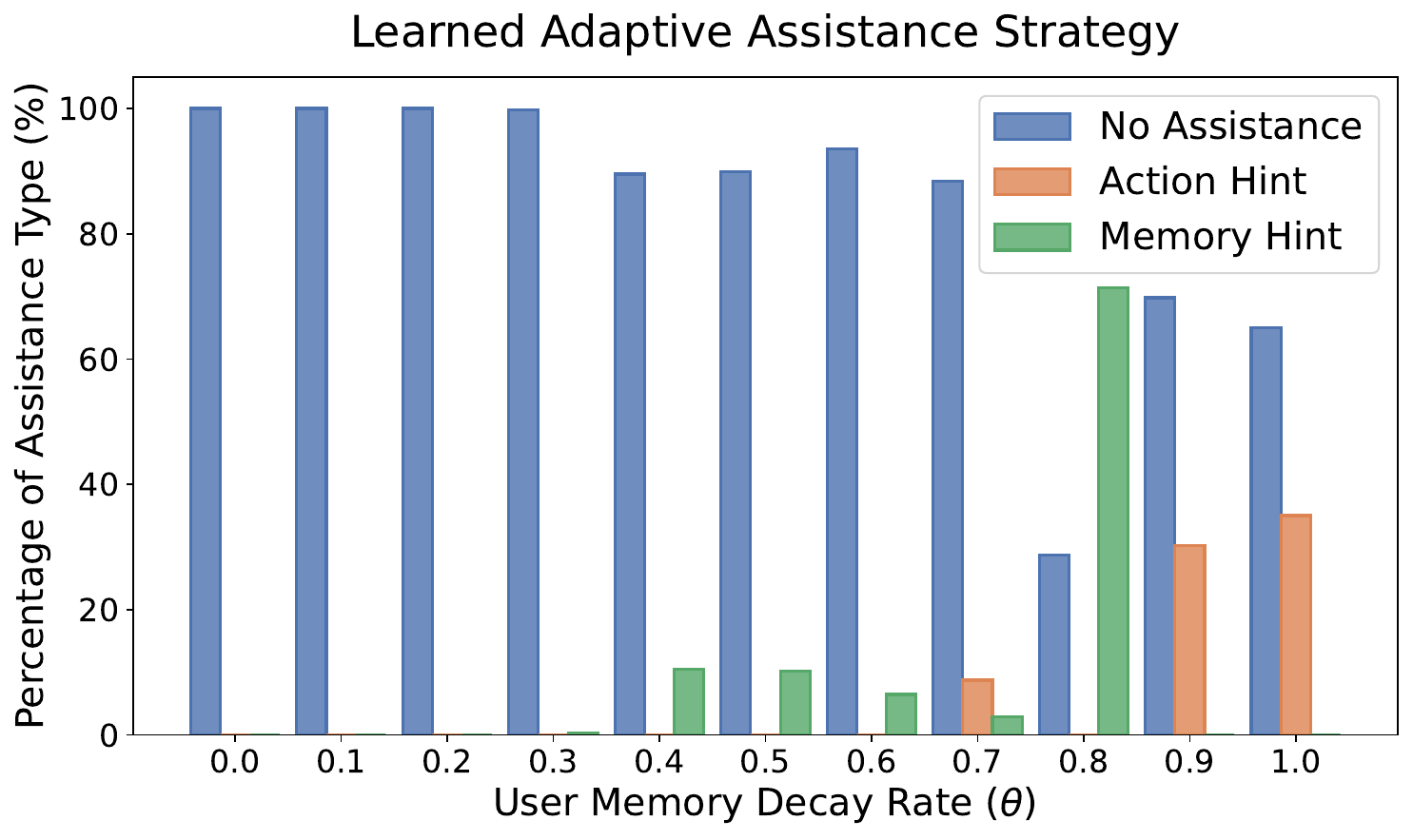}
\caption{
The learned adaptive assistance policy shows the distribution of assistance types provided by the AI assistant to simulated users with different ground-truth memory decay rates ($\theta$). The AI learned not to intervene for users with good memory ($\theta \le 0.3$), provide timely memory hints for moderately forgetful users, and provide more direct action hints for users with severe memory decay ($\theta \ge 0.9$).}
\label{fig:assistance_strategy}
\end{figure}

Together, the qualitative visualization of the posterior evolution in Figure \ref{fig:posterior_evolution} and the quantitative error analysis in Figure \ref{fig:convergence} provide strong empirical evidence in support of our central claim. 
They demonstrate that the distinct behavioral patterns generated by our CR model contain sufficient information for inference, and that our NPF-based method can effectively utilize this information, which supports our claim that the user's latent cognitive bound is practically identifiable within our proposed framework.

\subsection{Application Demonstration: Assistive-POMDP}
\subsubsection{Setup}
To demonstrate how our CR user model and inference method can benefit developing adaptive AI assistants, we designed an AI assistant aimed at maximizing the collaborative return while minimizing intervention cost. 
Here, the assistant observes the CR agent's actions and the task environment while the CR agent solves the navigation task, and is tasked with helping it. The AI is rewarded when the CR agent finds the goal.
The AI can do nothing or, for a small cost, remind the CR agent of a previous critical observation or, for a higher cost, suggest an action directly.
The AI is assumed to have access to the state of the environment, and thus knows where the goal is, but does not know the internal state $\tilde{h}$ or cognitive bounds $\theta$ of the user.
We formalize the AI's problem as a POMDP, where the internal state and parameters of the CR agent are hidden.
We solve this POMDP by using our proposed inference method to maintain a belief over those hidden quantities, and optimize the belief-based policy with PPO.
The detailed formalization of this assistive-POMDP framework and experiment setup, as well as the assistance policy learning algorithm, can be found in the extended version from our code repository.
\begin{figure}[t]
\centering
\includegraphics[width=0.88\columnwidth]{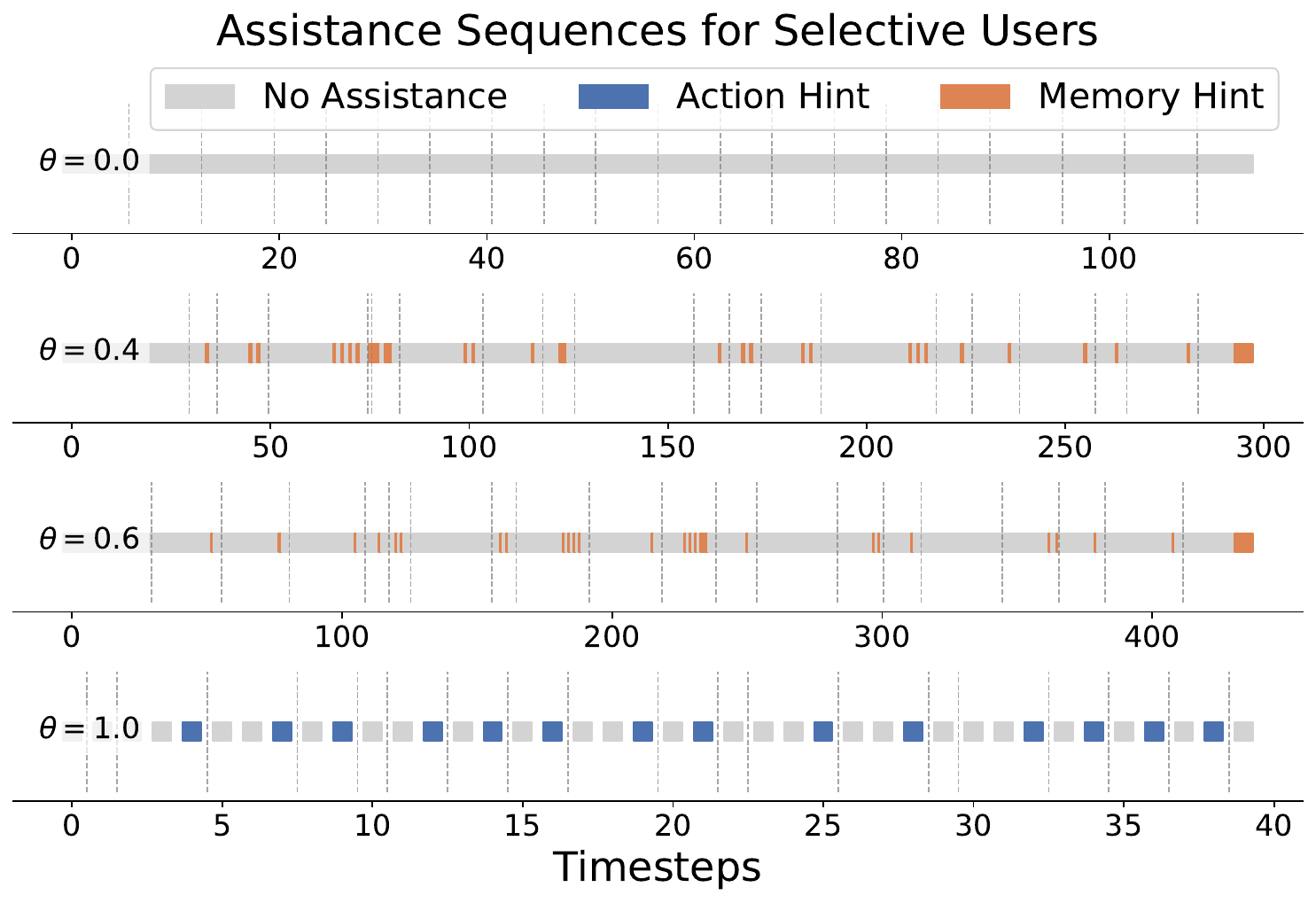}
\caption{The sequences show the timing of the learned assistance. Each panel shows the assistance sequence over multiple episodes for a user with memory bound $\theta$; The vertical dashed lines indicate episode boundaries. For moderately forgetful users ($\theta\in \{0.4,0.6\}$), memory hints are occasionally provided around the end of episodes, right before making the critical decision; For users with no memory $\theta=1.0$, only action hint is provided on the last step. }
\label{fig:assistance_sequences}
\end{figure}
\subsubsection{Results}
We evaluated our solution on simulated users with various memory bounds. 
The results demonstrate that the AI learned a highly adaptive policy that tailors both the type and timing of interventions to the inferred user cognitive bounds.
Figure \ref{fig:assistance_strategy} summarizes the learned policy by showing the distribution of each assistance type across the full spectrum of users, showing an intuitively adaptive strategy:
\begin{itemize}
    \item For users with low memory decay ($\theta \le 0.3$), the AI correctly infers that the user is memory competent and learned to rarely intervene. 
    \item For users with moderate memory decay ($\theta \in [0.4, 0.8]$), the AI identifies their need for cognitive support and provides significantly more memory hints and action hints.
    \item For users with severe memory decay ($\theta \ge 0.9$), the AI learns that simple memory hints are insufficient as the user is likely to forget them immediately, thus providing more direct action hints at critical moments.
\end{itemize}
To investigate the assistance timing, we plot an intervention sequence of a representative run in Figure \ref{fig:assistance_sequences}, which shows that the AI provides memory hints for moderately forgetful users ($\theta\in \{0.4,0.6\}$), and action hints for severely forgetful users ($\theta=1.0$) at the end of each episode and at moments that are critical or easy-to-forget. 
This demonstrates that the AI learned to intervene at moments of maximum utility, and that our online inference method provides a \emph{principled} foundation for learning an adaptive assistance policy.

\section{Discussion and Conclusion}\label{sec-discussion}
We have proposed a model, based on the concept of computational rationality, to explain irrational behavior as a result of imperfect memory.
Crucially, this model is flexible in its choice of memory model $f_{\theta}$, and allows for flexibility in the form of latent variables.
As part of our contribution, we also described an efficient inference algorithm based on nested particle filtering, demonstrating the effectiveness and interpretability of the model and inference technique in a non-trivial domain that reflects critical decision-making, with both qualitative and quantitative analysis.

There are limitations that, when addressed in future work, would further strengthen this research direction.
We want to highlight two of those:
First, for simplicity, we have assumed that both the user model and inference method have access to the underlying dynamics of the problem.
Second, while our choice of memory model in the experiments is reasonable, it is likely insufficient to capture realistically sophisticated agents.
Nevertheless, we believe this work is an important step toward a \emph{general} model for prediction and inference over the irrational behavior of agents due to imperfect memory.

\section{Acknowledgments}
This work was supported by the Research Council of Finland (Flagship programme: Finnish Center for Artificial Intelligence FCAI, Grant 359207), ELISE Networks of Excellence Centres (EU Horizon:2020 grant agreement 951847), and UKRI Turing AI World-Leading Researcher Fellowship (EP/W002973/1). 
We acknowledge the research environment provided by ELLIS Institute Finland. 
We also acknowledge the computational resources provided by the Aalto Science-IT Project from Computer Science IT and CSC–IT Center for Science, Finland.

\bibliography{aaai2026}

\appendix

\maketitle
\begin{figure*}[p]
    \centering 
    \begin{subfigure}[b]{0.24\linewidth}
        \centering
        \includegraphics[width=\linewidth]{imgs/trajectory_theta_0.0_seed_66_temp_1.png}
        \caption{Perfect Memory ($\theta=0.0$)}
        \label{fig:traj_perfectmem}
    \end{subfigure}
    \hfill 
    \begin{subfigure}[b]{0.24\linewidth}
    \centering
        \includegraphics[width=\linewidth]{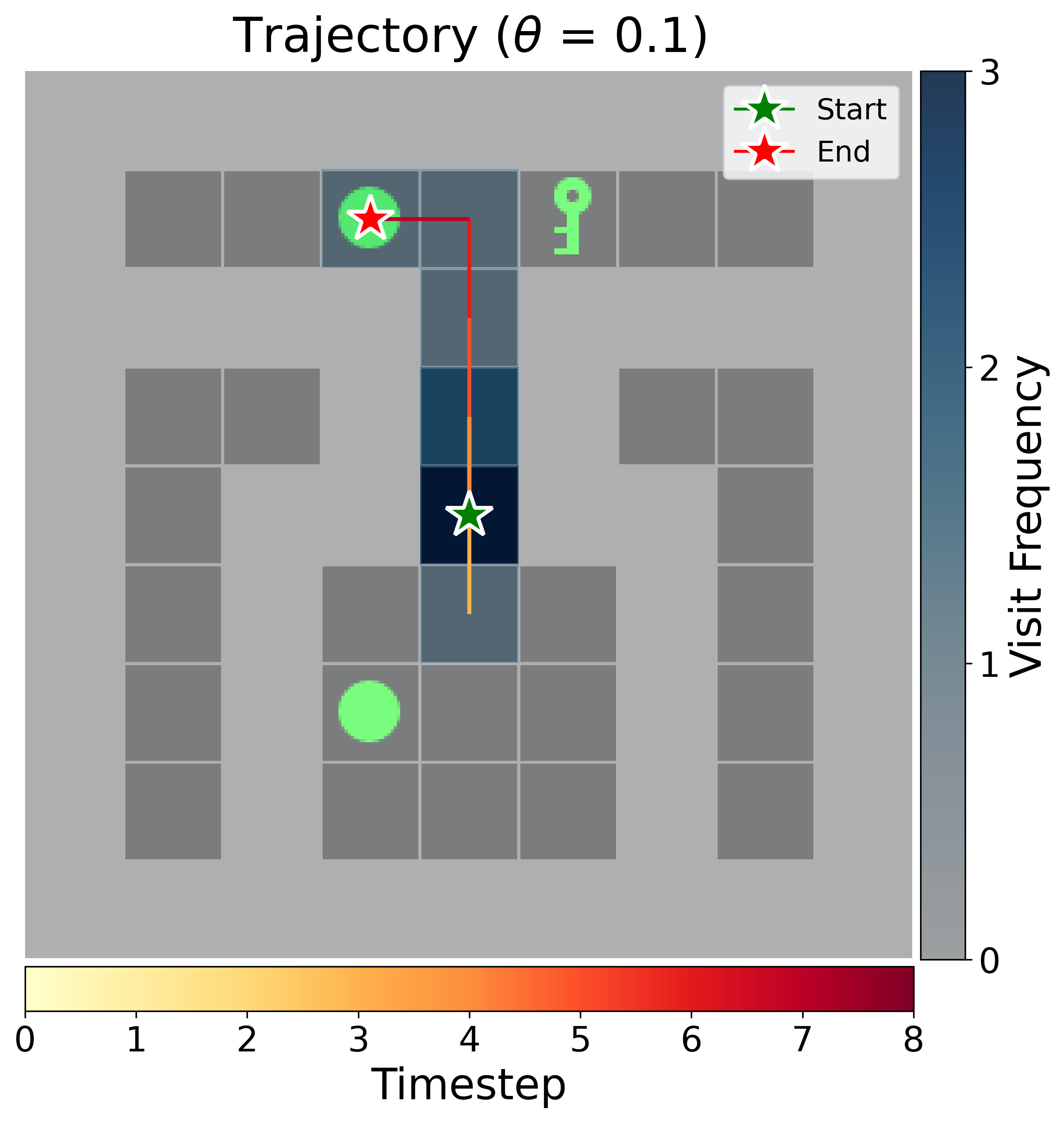}
        \caption{Imperfect Memory ($\theta=0.1$)}
        \label{fig:traj_imperfect_0.1}
    \end{subfigure}
    \hfill 
    \begin{subfigure}[b]{0.24\linewidth}
    \centering
        \includegraphics[width=\linewidth]{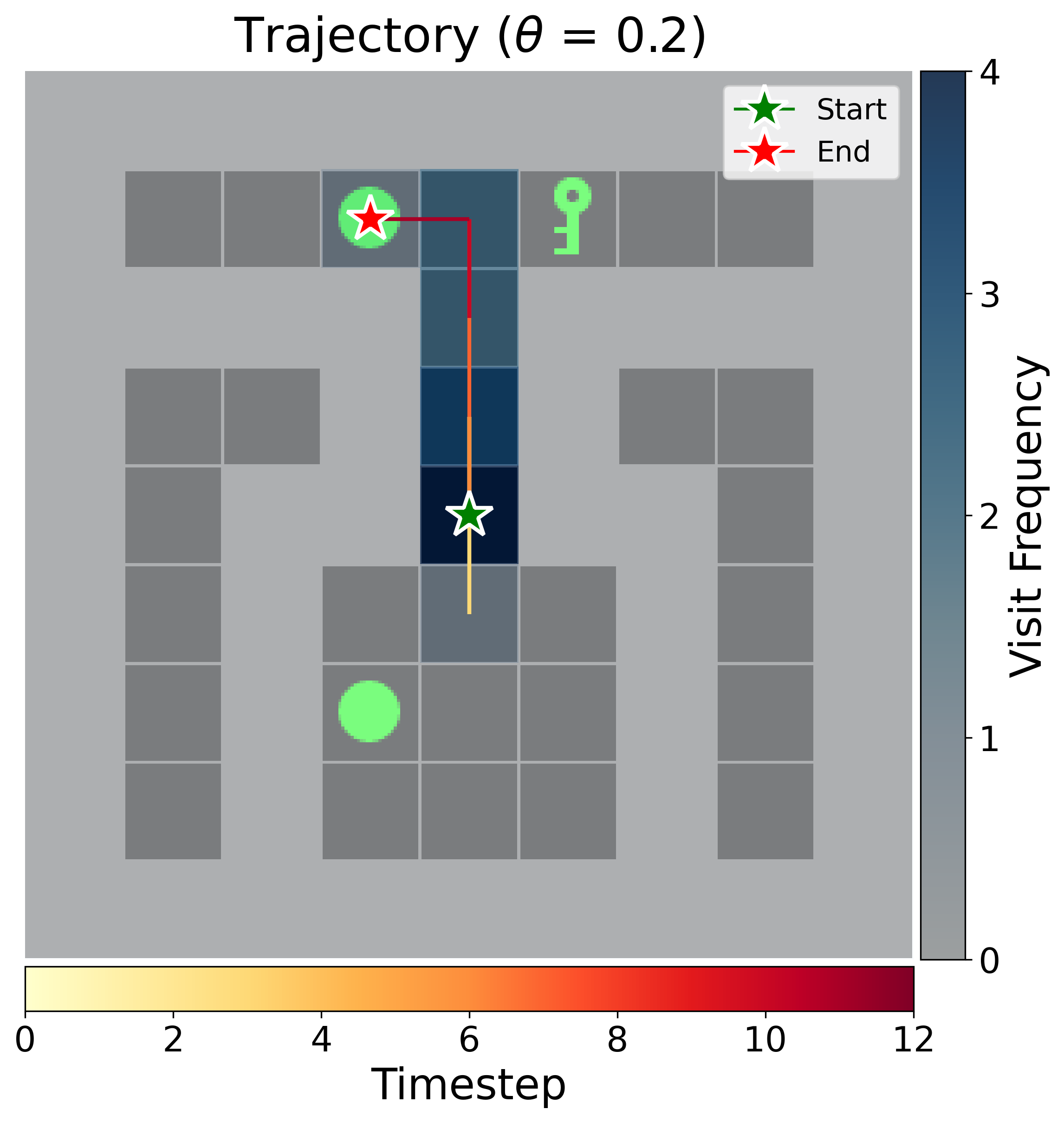}
        \caption{Imperfect Memory ($\theta=0.2$)}
        \label{fig:traj_imperfect_0.2}
    \end{subfigure}
    \hfill 
    \begin{subfigure}[b]{0.24\linewidth}
    \centering
        \includegraphics[width=\linewidth]{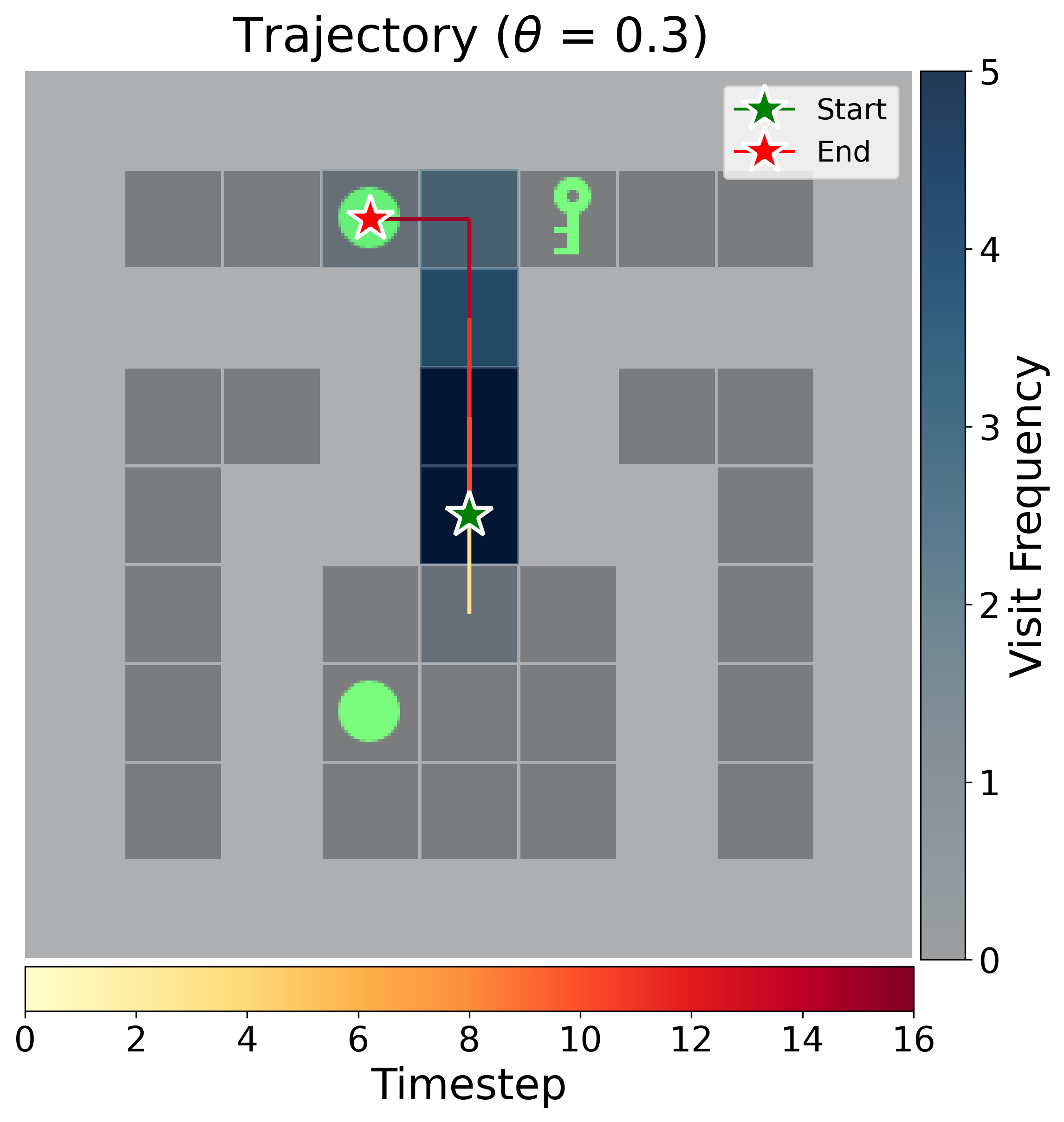}
        \caption{Imperfect Memory ($\theta=0.3$)}
        \label{fig:traj_imperfect_0.3}
    \end{subfigure}
    \hfill 
    \begin{subfigure}[b]{0.24\linewidth}
    \centering
        \includegraphics[width=\linewidth]{imgs/trajectory_theta_0.4_seed_66_temp_1.png}
        \caption{Imperfect Memory ($\theta=0.4$)}
        \label{fig:traj_imperfect_0.4}
    \end{subfigure}
    \hfill 
    \begin{subfigure}[b]{0.24\linewidth}
    \centering
        \includegraphics[width=\linewidth]{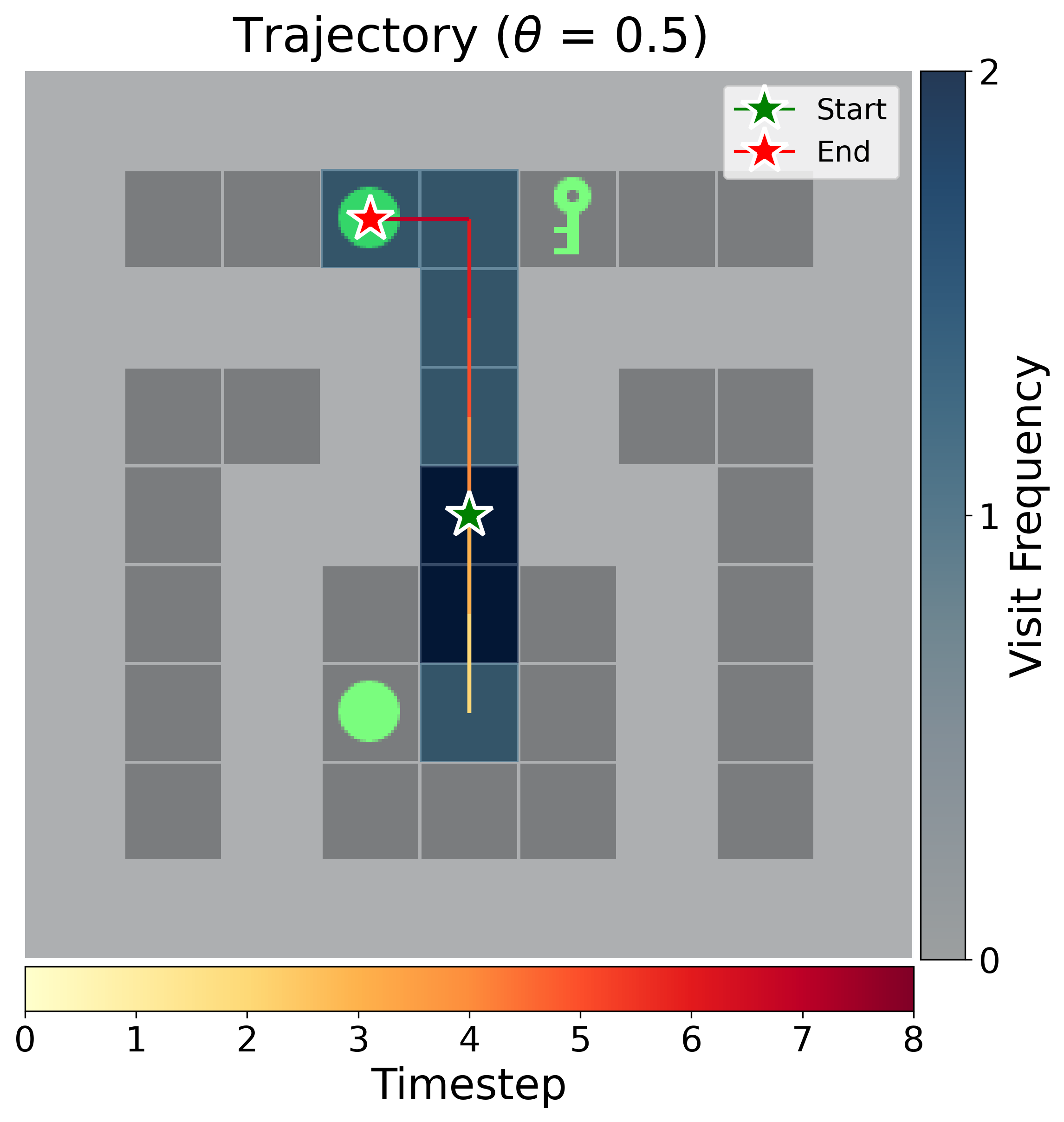}
        \caption{Imperfect Memory ($\theta=0.5$)}
        \label{fig:traj_imperfect_0.5}
    \end{subfigure}
    \hfill 
    \begin{subfigure}[b]{0.24\linewidth}
    \centering
        \includegraphics[width=\linewidth]{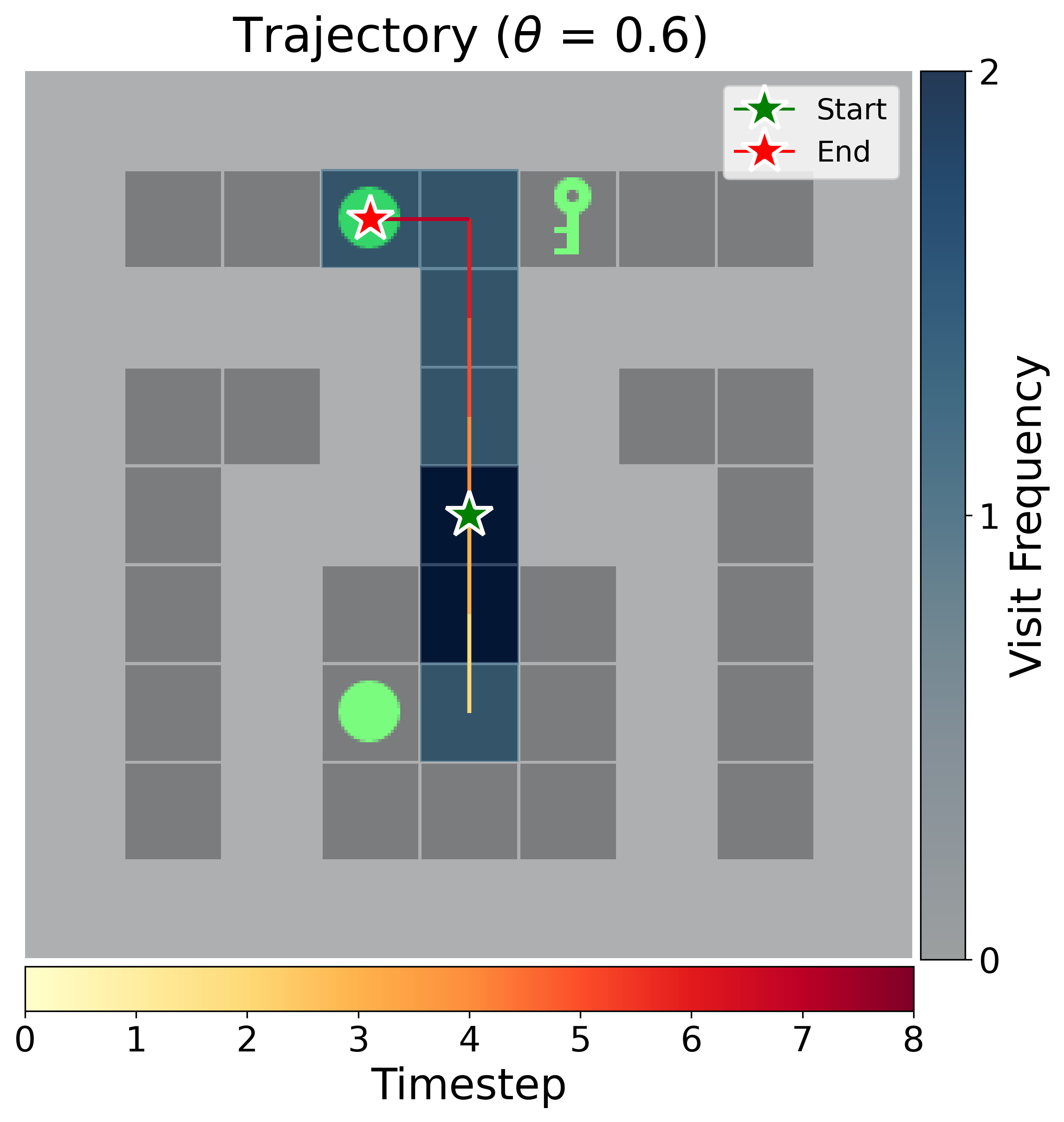}
        \caption{Imperfect Memory ($\theta=0.6$)}
        \label{fig:traj_imperfect_0.6}
    \end{subfigure}
    \hfill 
    \begin{subfigure}[b]{0.24\linewidth}
    \centering
        \includegraphics[width=\linewidth]{imgs/trajectory_theta_0.7_seed_66_temp_1.png}
        \caption{Imperfect Memory ($\theta=0.7$)}
        \label{fig:traj_imperfect_0.7}
    \end{subfigure}
    \hfill 
    \begin{subfigure}[b]{0.24\linewidth}
    \centering
        \includegraphics[width=\linewidth]{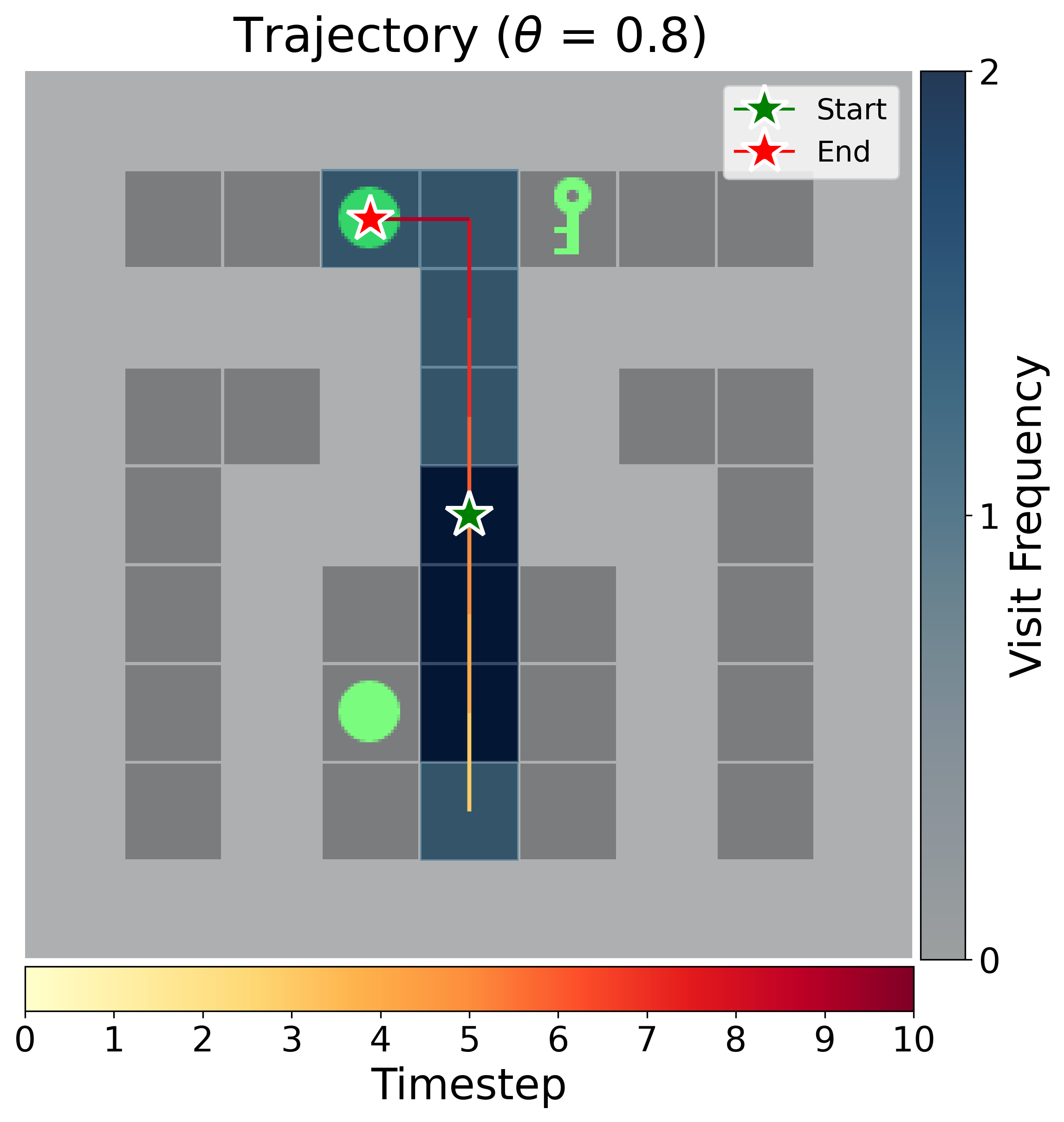}
        \caption{Imperfect Memory ($\theta=0.8$)}
        \label{fig:traj_imperfect_0.8}
    \end{subfigure}
    \hfill 
    \begin{subfigure}[b]{0.24\linewidth}
    \centering
        \includegraphics[width=\linewidth]{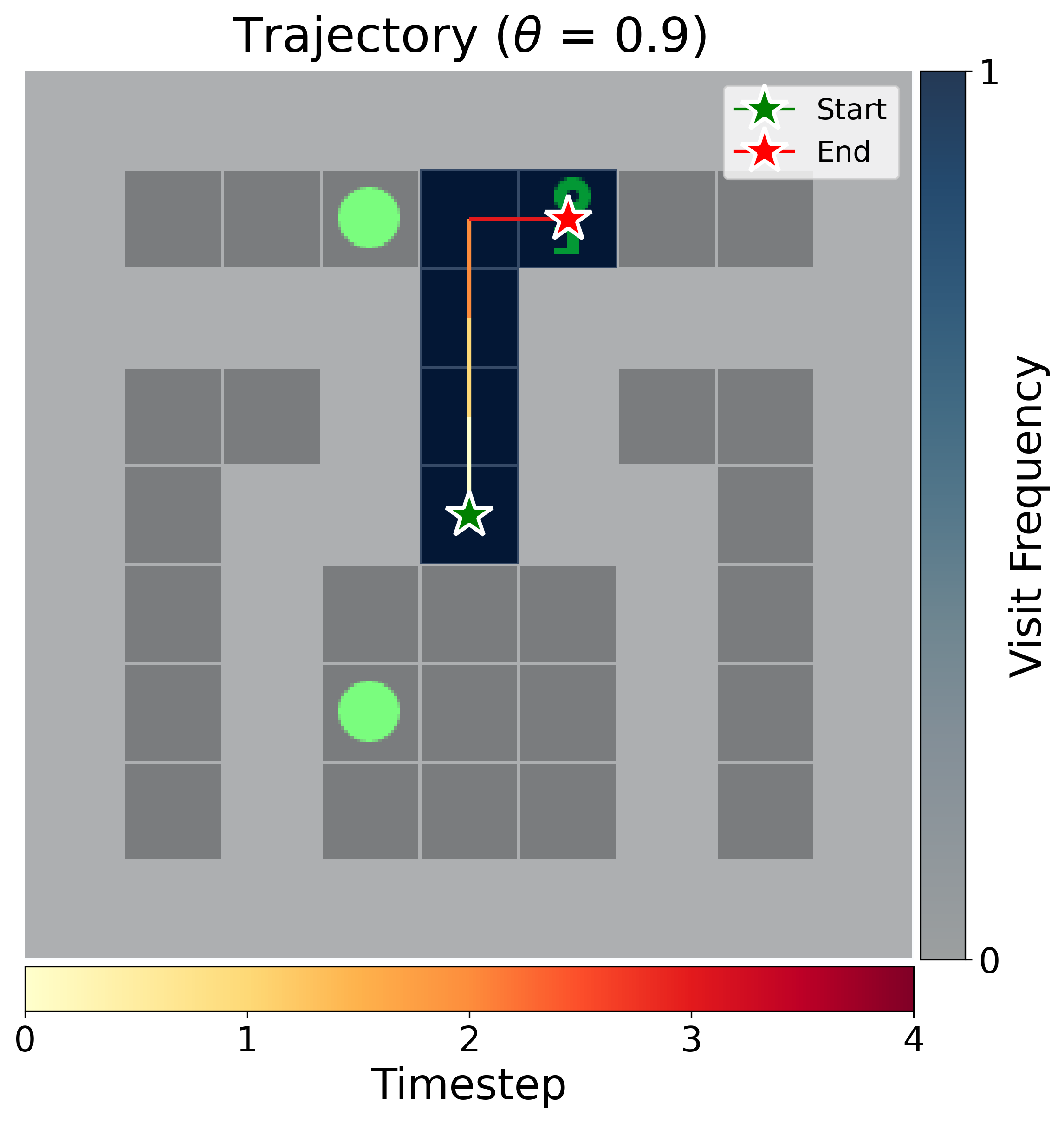}
        \caption{Imperfect Memory ($\theta=0.9$)}
        \label{fig:traj_imperfect_0.9}
    \end{subfigure}
    \hfill 
    \begin{subfigure}[b]{0.24\linewidth}
    \centering
        \includegraphics[width=\linewidth]{imgs/trajectory_theta_1.0_seed_66_temp_1.png}
        \caption{No Memory ($\theta=1.0$)}
        \label{fig:traj_nomem}
    \end{subfigure}
    \caption{Representative trajectories generated by our CR model for agents with different memory bound parameters $\theta$. 
    The color of the trajectory indicates the temporal order of the steps (brighter is earlier), and the colored grids indicate the visit frequency.
    The agent starts in the hallway (green star) and reaches one of the terminal states (red star). (a) With perfect memory, the agent acts optimally by going down to find the object and directly proceeds to the correct terminal state. (b-j) With imperfect memory, the agent shows different behavioral patterns according to different memory decay rates. For example, with a 40\% chance of losing memory, the agent learned to collect more observations on the object; With a 70\% chance of losing memory, the agent learned to recheck the found position after it appeared to have forgotten. (k) With no memory, the agent learned to take a random guess without exploring the environment. }
    \label{fig:app_exp_crtraj}
\end{figure*}
\appendix

\section{Additional Details for CR User Model Validation}
To evaluate the behavior trajectories generated by our CR user model in the T-maze simulation task, we pre-trained 11 optimal policies $\pi_*(a\mid \tilde{b};\theta)$ for CR agents with memory bound $\theta \in \{0.0, 0.1,...,1.0\}$ using Proximal Policy Optimization (PPO). 
Figure \ref{fig:app_exp_crtraj} presents the illustrations of trajectories from all pre-trained CR policies, which show that with imperfect
memory, the agent shows different behavioral patterns according to different memory decay rates.
\section{Additional Details for Online Inference Validation}
\subsection{Hyperparameter Sensitivity Analysis}
\begin{figure}[h!]
\centering
\includegraphics[width=\columnwidth]{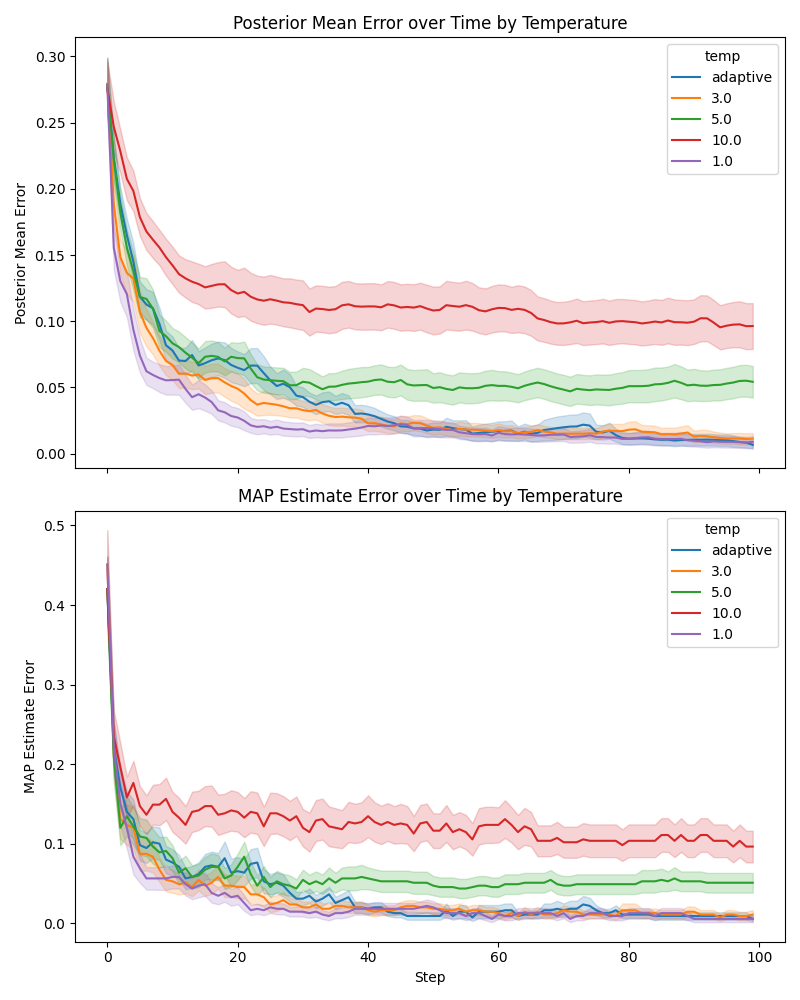}
\caption{Sensitivity analysis for the softmax temperature $\tau$. The plots show the convergence of PM and MAP errors for different temperature settings. Our method demonstrates robust performance, successfully converging in all but the noisiest ($\tau=10.0$) conditions.}
\label{fig:sensitivity_temp}
\end{figure}

\paragraph{Sensitivity to Softmax Temperature ($\tau$).}
The temperature parameter, $\tau$, in the softmax function (used to simulate the CR agent's actions during online inference) controls the level of stochasticity in the agent's behavior. A high $\tau$ leads to more random actions, while a low $\tau$ leads to more deterministic (optimal) actions. This parameter directly affects the quality of the behavioral evidence available to the AI assistant for inference. To test the robustness of our NPF-based method, we evaluated its performance across a range of temperature settings.
We tested four fixed temperature values, $\tau \in \{1.0, 3.0, 5.0, 10.0\}$ and one adaptive schedule. The adaptive temperature starts at $\tau_0=5.0$ and exponentially decays to $\tau_T=1.0$ at step $100$, simulating an agent that becomes more confident over time.
Figure \ref{fig:sensitivity_temp} shows the convergence of the posterior mean (PM) Error and the MAP Estimate Error for each temperature setting, aggregated over all ground-truth $\theta$ conditions and 5 random seeds. The results demonstrate the robustness of our inference method:
\begin{itemize}
\item Across all tested conditions (adaptive, $\tau=1.0$, $\tau=3.0$, $\tau=5.0$), our method successfully converges to a low error, confirming that it can accurately recover the latent bound even under varying levels of behavioral noise.
\item As expected, a very high temperature ($\tau=10.0$, red line) leads to slower convergence and higher final error. This is because highly stochastic actions provide less identifiable information about the agent's underlying cognitive bounds.
\item The adaptive temperature schedule and the lower fixed temperatures ($\tau \in \{1.0, 3.0, 5.0\}$) yield the best and most stable performance. 
\end{itemize}
While the adaptive temperature shows good performance, a fixed temperature is a more cognitively plausible assumption for an agent with a stable level of decision noise.
For the main paper, we chose to report the results from the $\tau=3.0$ setting as it represents a competent agent with a reasonable level of stochasticity.
\section{Additional Details for Assistive-POMDP}
Here we present the design of our adaptive AI assistant for CR users. 
We begin by formulating the AI-assisted decision-making process for CR agents as an assistive-POMDP problem, where the goal is to maximize collaboration utility while minimizing intervention cost.
Achieving this goal requires the assistant to understand user's sub-optimal behaviors.
We formulate the understanding problem as an online Bayesian inference problem on estimating both user's belief state and cognitive bounds, and address it using nested particle filtering.
Finally, we introduce an online adaptive assistance strategy, which explicitly leverages the real-time estimates to make context-aware decisions about when and how to provide assistance, thereby closing the loop from understanding the user to providing adaptive assistance.
\subsection{The Assistive-POMDP: AI-Assisted Decision Making Framework}
We formalize AI assistant's decision-making process as an assistive-POMDP where the environment consists of a CR agent and a shared task environment. Here we assume that the AI can fully observe the task environment but can only partially observe the CR agent, meaning that the AI has no access to the CR agent's cognitive bounds and belief state but they can observe their actions and the true environment states. 

We define the assistive-POMDP as a tuple $\mathcal{M}^{\texttt{AI}} = (\mathcal{S}^{\texttt{AI}}, \mathcal{A}^{\texttt{AI}}, \mathcal{T}^{\texttt{AI}}, \mathcal{R}^{\texttt{AI}}, \Omega^{\texttt{AI}}, \mathcal{O}^{\texttt{AI}},\gamma^{\texttt{AI}})$, and refer to the decision-making process of the CR agent as $\mathcal{M}^{\texttt{H}}_{\theta} = (\mathcal{S}, \mathcal{A}^{\texttt{H}}, \mathcal{T}, \mathcal{R}, \Omega, \mathcal{O},\gamma^\texttt{H})$. We first present our definition for each component in this AI-assisted decision-making framework, and then describe the interaction dynamics. 
\paragraph{State space $\mathcal{S}^{\texttt{AI}}$} includes both environment states $s_t \in \mathcal{S}$ and the CR agent's cognitive state $\tilde{h}_{t-1} = (\tilde{\mathbf{o}}_{t-1}^{:t-1}, \tilde{\mathbf{a}}_{t-1}^{:t-2})$ as well as cognitive bounds $\theta \in \Theta$. 
Thus, $s_t^{\texttt{AI}} = (s_t, \tilde{h}_{t-1}, \theta)$, where the environment state $s_t$ is known while agent's cognitive state $\tilde{h}_{t-1}$ as well as bounds $\theta$ remain latent. Here we define the agent's cognitive state in $s_t^{\texttt{AI}}$ to be $\tilde{h}_{t-1}$ and not $\tilde{h}_t$ because $\tilde{h}_{t-1}$ is determined at time $t$ for the AI while $\tilde{h}_t$ depends on the AI's assistance $a_{t}^\texttt{AI}$. The AI thus needs to maintain beliefs over agent's unknown states via inferring $(\tilde{h}_{t-1}, \theta)$. 
\paragraph{Action space $\mathcal{A}^{\texttt{AI}}$} is a finite set of actions designed for providing various levels of assistance, designed as $\mathcal{A}^{\texttt{AI}} = \{\texttt{DoNothing}, \texttt{ActionHint}, \texttt{MemoryHint}\}$. The assistant can choose $a_{\texttt{DN}}^{\texttt{AI}}$ (\texttt{DoNothing}) to not intervene the agent, $a_{\texttt{AH}}^{\texttt{\texttt{AI}}} \in \mathcal{A}^{\texttt{H}}$ (\texttt{ActionHint}) to suggest the current optimal action given the environment state and the optimal MDP policy $a_{\texttt{AH}}^{\texttt{\texttt{AI}}} = \arg\max_a \pi_*(\cdot\mid s_t)$, or $a_{\texttt{MH}}^{\texttt{AI}}$ (\texttt{MemoryHint}) to remind the agent of an critical observation seen earlier but may have been forgotten. Here we define $a_{\texttt{MH}}^{\texttt{AI}} = (k,o_k)$, where $o_k \in \Omega$ is the observation received from the environment at time $k$. 
\paragraph{Agent reaction}\label{sec-agent-reaction} refers to the effect of AI assistance $a^{\texttt{AI}}_t$ provided at time $t$ on the CR agent's decision-making process. Here we define how $a^{\texttt{AI}}_t$ affects the agent's cognitive state $\tilde{h}_t$, belief state $b_t^\texttt{H}(s_t;\theta)$, and action selection $a_t^{\texttt{H}}$:
\begin{enumerate}
    \item No intervention ($a_t^{\texttt{AI}} = a_{\texttt{DN}}^{\texttt{AI}}$): no effect is made on the agent when the AI stays quiet.
    \item Action hint ($a_t^{\texttt{AI}} = a_{\texttt{AH}}^{\texttt{AI}}$): when the AI provides an action suggestion, the agent accepts $a_t^{\texttt{AI}}$ as per current decision uncertainty, while current cognitive state and belief are not directly affected. We quantify the agent's decision uncertainty with the entropy of its policy: $H_t(\pi^{\texttt{H}})=-\sum_{a^\texttt{H}\in \mathcal{A}^\texttt{H}}\pi^\texttt{H}(a^\texttt{H} \mid b_t^\texttt{H};\theta)\log \pi^\texttt{H}(a^\texttt{H} \mid b_t^\texttt{H};\theta)$. Let $H$ denote a pre-defined uncertainty threshold, we model the agent's assisted policy as follows:
    \begin{equation}
        a_t^\texttt{H} = 
        \begin{cases}
            a_t^{\texttt{AI}} & \text{if }H_t(\pi^{\texttt{H}}) \ge H\\
            a \sim \pi^\texttt{H}(a^\texttt{H}\mid b_t^\texttt{H};\theta) & \text{otherwise}.
        \end{cases}
    \end{equation}
    The agent accepts AI's action suggestion when they are highly uncertain (high entropy), and otherwise sticks to their own policy. 
    \item Memory hint $a^{\texttt{AI}} = a_{\texttt{MH}}^{\texttt{AI}}$: when the AI provides a memory hint $(k, o_k)$, which can be seen as an auxiliary observation, the agent always trust the hint and uses it to refresh the specific memory content. Specifically, the agent first update their cognitive state to process the new environment observation and update memory: $\tilde{h}_t  \sim f_\theta(\tilde{h}_{t-1}, o_t)$; Then refresh their specific memory as per the hint:
    \begin{equation}\label{eq-ai-memoryhint}
    \widetilde{o}_i^t = 
        \begin{cases} 
            o_k & \text{if } i = k \\
            \widetilde{o}_i^{t} & \text{otherwise.}
        \end{cases}
    \end{equation}
    This reaction design explicitly models the refreshing of the agent's potentially faded memory $\widetilde{o}_k^t$ with the memory hint. 
    The agent then computes $b_t^\texttt{H}(s_t;\theta)$ with the updated $\tilde{h}_t$ and selects actions accordingly $a_t^\texttt{H} \sim \pi^\texttt{H}(\cdot \mid b_t^\texttt{H};\theta)$. 
\end{enumerate}
Thus, the agent's policy under assistance $p(a_t^\texttt{H} \mid b_t^\texttt{H}, a_t^\texttt{AI};\theta)$ is formalized as 
\begin{equation}\label{eq-agent-assistedpolicy}
     a \sim 
    \begin{cases}
        \delta(a_t^\texttt{H} - a_t^\texttt{AI}) &\text{if $a_t^\texttt{AI} = a_{\texttt{AH}}^{\texttt{AI}}$ and $H_t(\pi^\texttt{H}) \ge H$}\\
        \pi^\texttt{H}(a_t^\texttt{H} \mid b_t^\texttt{H};\theta) & \text{otherwise.}
    \end{cases}
\end{equation}

We propose this agent reaction model to capture key characteristics of human decision-making with assistance. The intuition behind this design is that humans tend to accept direct action guidance (\texttt{ActionHint}) when uncertain, but rely on their own judgement when confident. For $\texttt{MemoryHint}$, humans tend to believe and incorporate a reminded fact received from a reliable system, such as AI. This reaction model is critical for defining AI's transition dynamics below. 
\paragraph{Transition dynamics $\mathcal{T}^{\texttt{AI}}$} defines the evolution of AI's states $s_t^\texttt{AI} = (s_t, \tilde{h}_{t-1},\theta)$. Here we assume that only the agent's action $a_t^\texttt{H}$ impacts the environment states $s_t$ via $\mathcal{T}(s_{t+1}\mid s_t,a_t^\texttt{H})$, and that the agent's cognitive bounds $\theta$ stay stationary. Thus, $\mathcal{T}^{\texttt{AI}}(s_{t+1}^\texttt{AI}\mid s_t^\texttt{AI}, a_t^{\texttt{AI}})$ consists of: (i) environment transition dynamics $\mathcal{T}(s_{t+1}\mid s_t,a_t^{\texttt{H}})$, and (ii) agent's cognitive transition dynamics $p(\tilde{h}_t \mid \tilde{h}_{t-1}, o_{t}, a_{t}^\texttt{AI}, \theta)$. Specifically, for the latter, if $a^{\texttt{AI}}_t = a^{\texttt{AI}}_{\texttt{MH}}$, then $\tilde{h}_t$ is updated and refreshed by Equation \eqref{eq-ai-memoryhint}; Otherwise, $\tilde{h}_t$ is directly updated by $f_{\theta}(\tilde{h}_{t-1}, o_{t})$:
\begin{equation}\label{eq-ai-cognitive-dynamics}
    \tilde{h}_t \sim 
    \begin{cases}
        f_{\theta}(\tilde{h}_{t-1}, o_t), \widetilde{o}_k^t=o_k &\text{if $a_{t}^\texttt{AI} = a^\texttt{AI}_{\texttt{MH}}(k, o_k)$} \\
        f_{\theta}(\tilde{h}_{t-1}, o_t) &\text{otherwise}
    \end{cases}
\end{equation}
\paragraph{Observation space $\Omega^{\texttt{AI}}$} is defined as $\Omega \times \mathcal{A}^\texttt{H}$, the Cartesian product of the environment observation space and the agent's action space. Here we assume that the AI observes the previous action $a_{t-1}^\texttt{H} \in \mathcal{A}^\texttt{H}$ taken by the agent and received the same observation $o_t \in \Omega$ as the agent from the environment, and thus $o_t^{\texttt{AI}} = (o_t, a_{t-1}^\texttt{H})$. 
\paragraph{Observation function $\mathcal{O}^{\texttt{AI}}$} specifies the probability of the AI observing the specific pair $o_t^{\texttt{AI}} = (o_t, a_{t-1}^{\texttt{H}})$ given the underlying state $s_t^{\texttt{AI}} = (s_t, \tilde{h}_{t-1}, \theta)$ and the AI's previous action $a_{t-1}^{\texttt{AI}}$: 
\begin{equation}\label{eq-ai-obs}
    \begin{aligned}
        &p(o_t, a_{t-1}^\texttt{H}\mid s_t^\texttt{AI}, a_{t-1}^\texttt{AI}) = \\ 
        & \qquad \qquad\underbrace{p(a_{t-1}^\texttt{H}\mid \tilde{h}_{t-1}, \theta,  a_{t-1}^{\texttt{AI}})}_{\text{Agent Assisted Policy}}\mathcal{O}(o_t\mid s_t)
    \end{aligned}
\end{equation}
where the agent's policy under assistance is defined by Equation \eqref{eq-agent-assistedpolicy}.
\paragraph{Belief $b^{\texttt{AI}}$} refers to AI's inference on the agent's latent cognitive state and the fixed cognitive bounds, defined as  $b_t^\texttt{AI} = p(\theta, \tilde{h}_{t-1}\mid h_t)$, where $h_t = (s_{:t}, a_{:t-1}^{\texttt{AI}}, o_{:t}^{\texttt{AI}})$ is the interaction history up to time $t$, consisting of environment states, AI assistance, and AI observations. $b_t^\texttt{AI}$ is computed by
\begin{equation}\label{eq-aidm-belief}
\begin{aligned}
    &b_t^\texttt{AI}(\tilde{h}_{t-1},\theta) = \alpha \mathcal{O}^\texttt{AI}(o_t^\texttt{AI}\mid s_t^\texttt{AI},a_{t-1}^\texttt{AI}) \\
    & \quad \times\sum_{\tilde{h}_{t-2}}\mathcal{T}^\texttt{AI}(s_t^\texttt{AI}\mid s_{t-1}^\texttt{AI}, a_{t-1}^\texttt{AI}, a_{t-1}^\texttt{H})b_{t-1}^\texttt{AI}(\tilde{h}_{t-2},\theta)
\end{aligned}
\end{equation}
where $\alpha$ is a normalization constant. Providing adaptive assistance online requires the AI to maintain accurate beliefs on the agent's current cognitive state and bound, enabling predicting the agent's future actions and evaluating the assistance. However, the belief computation is intractable because (i) the latent space grows over time and can be high-dimensional; (ii) The observation likelihood and transition probabilities computation requires simulating the agent's complex internal cognitive dynamics, which is computationally intensive. We propose a solution based on nested particle filtering to approximate AI's belief. 

\paragraph{Reward function $\mathcal{R}^{\texttt{AI}}$} corresponds to the AI's objective in this AI-assisted decision-making setting, which is to maximize collaboration utility while minimizing intervention cost. We formulate the immediate reward received by the AI at time $t$ as $r^{\texttt{AI}}(s_t, a_t^\texttt{H}, a_t^{\texttt{AI}}) = r(s_t, a_t^\texttt{H}) - c(a_t^\texttt{AI})$, where $r(s_t, a_t^\texttt{H})$ is the task reward and $c(a_t^\texttt{AI})$ is the intervention cost. Accounting for the agent's cognitive load and the computation cost of generating the assistance, we define a cost function $c(a^\texttt{AI})$ for AI assistance:
\begin{itemize}
    \item $c(a_{\texttt{DN}}^{\texttt{AI}}) = 0$: no cost when the AI stays quiet.
    \item $c(a_{\texttt{AH}}^{\texttt{AI}}) = c_{\texttt{AH}}$: a positive cost $c_{\texttt{AH}} \in \mathbb{R}^+$ is assigned when the AI provides an action hint.
    \item $c(a_{\texttt{MH}}^{\texttt{AI}}) = c_{\texttt{MH}}$: a positive cost $c_{\texttt{MH}} \in \mathbb{R}^+$ is assigned when the AI provides a memory hint.
\end{itemize}
The AI's goal is to learn a policy $\pi^{\texttt{AI}}$ that maximizes the expected cumulative reward $\mathbb{E}[\sum_{t=0}^T (\gamma^{\texttt{AI}})^t r^{\texttt{AI}}(s_t, a_t^{\texttt{H}}, a_t^{\texttt{AI}})]$. 

\paragraph{AI assistance $\pi^{\texttt{AI}}(a_t^\texttt{AI}\mid b_t^{\texttt{AI}}, s_t)$} is the AI's policy for assisting the agent. The fundamental problem the policy solves is twofold:
\begin{enumerate}
    \item Determining the optimal \emph{type} of interaction: Should the AI intervene (\texttt{ActionHint} or \texttt{MemoryHint}) or stay quiet (\texttt{DoNothing})?
    \item If intervening, determining the optimal \emph{content}: Which specific action $a \in \mathcal{A}$ should be suggested? Which specific past memory $(k, o_k)$ should be reminded?
\end{enumerate}
Learning an effective policy $\pi^{\texttt{AI}}$ poses significant computational challenges inherent to this assistive-POMDP formulation. Firstly, the policy must operate over the high-dimensional belief space $b_t^\texttt{AI}$, whose complexity arises from the sequential and potentially growing nature of the agent's cognitive state $\tilde{h}_{t-1}$. Standard POMDP planning and reinforcement learning algorithms often struggle with such large belief spaces. Secondly, the AI must select actions from the large, structured, and parameterized action space $\mathcal{A}^{\texttt{AI}}$, requiring efficient exploration and optimization strategies, especially for identifying the most relevant memory hint $(k, o_k)$ among numerous past experiences. Finally, evaluating the long-term expected return for any candidate action $a_t^{\texttt{AI}}$ necessitates potentially costly rollouts or simulations using the complex underlying transition ($\mathcal{T}^{\texttt{AI}}$) and observation ($\mathcal{O}^{\texttt{AI}}$) models, which embed the agent's intricate cognitive dynamics and reactions.

Addressing these challenges—high-dimensional beliefs, structured actions, and complex dynamics simulation for long-term evaluation—is crucial for realizing effective adaptive assistance. The specific methods we employ for learning the adaptive assistance policy $\pi^{\texttt{AI}}$ are detailed in the section below.

The AI-assisted decision-making process is described as follows. At time $t$, both the AI and the agent are in environment state $s_t$. The AI's state is $s_t^{\texttt{AI}} = (s_t, \tilde{h}_{t-1}, \theta)$, where $\tilde{h}_{t-1}$ is the agent's cognitive state at time $t-1$. The AI selects assistance according to its current belief $b_t^\texttt{AI}(\tilde{h}_{t-1},\theta)$ and $s_t$: $a_t^{\texttt{AI}}\sim \pi^\texttt{AI}(b_t^\texttt{AI}(\tilde{h}_{t-1},\theta), s_t)$. The agent receives environment observation $o_t$ as well as assistance $a_t^{\texttt{AI}}$, and then updates its cognitive state $\tilde{h}_t \sim f_{\theta}(\tilde{h}_{t-1}, o_t)$ according to Equation \eqref{eq-ai-cognitive-dynamics}. The agent then computes beliefs $b_t^\texttt{H}$, and selects action $a_t^\texttt{H}$ according to Equation \eqref{eq-agent-assistedpolicy}. The environment then evolves to the next state $s_{t+1}\sim \mathcal{T}(\cdot \mid s_t,a_t^\texttt{H})$, so as the AI's state $s_{t+1}^{\texttt{AI}}=(s_{t+1}, \tilde{h}_t, \theta) \sim \mathcal{T}^{\texttt{AI}}(\cdot \mid s_t^\texttt{AI}, a_t^\texttt{AI})$. The AI receives new observation $o_{t+1}^\texttt{AI} = (o_{t+1}, a^{\texttt{H}}_{t})$, and updates its belief over the agent's cognitive state and cognitive bounds with Equation \eqref{eq-aidm-belief}.

\subsection{Adaptive Assistance via Reinforcement Learning}\label{sec-assistance-learning}
The adaptive assistance policy should enable optimally deciding when and how to intervene the user given current belief $b_t^\texttt{AI}$ about the user's cognitive state and bounds. Given the computational challenges, here we first propose a hybrid approach for learning such policies. Specifically, we simplify the assisting decision-making $\pi^\texttt{AI}(a_t^\texttt{AI}\mid b_t^\texttt{AI},s_t)$ as a two-step process:
\begin{enumerate}
    \item Selecting assisting type via an RL policy $a_t^\texttt{AI-type} \sim \pi^\texttt{AI-type}_{\psi}(\cdot \mid b^\texttt{AI}_t)$ where $a_t^\texttt{AI-type} \in \{\texttt{DoNothing}, \texttt{ActionHint}, \texttt{MemoryHint}\}$.
    \item Selecting assisting content $a_t^\texttt{AI-content}$ as per $a_t^\texttt{AI-type}$ via a rule-based policy. If $a_t^\texttt{AI-type} = \texttt{ActionHint}$, then the optimal action computed from the known optimal MDP policy $a^* = \arg\max_{a\in \mathcal{A}^\texttt{H}}\pi_*(\cdot \mid s_t)$ is provided as $a_t^\texttt{AI-content}$; If $a_t^\texttt{AI-type} = \texttt{MemoryHint}$, then a critical observation $(k, o_k)$ is privided as $a_t^\texttt{AI-content}$, which is sampled from a list of critical observations saved upon evaluation from a pre-defined criterion based on prior knowledge about the task. 
\end{enumerate}
The RL policy $\pi^\texttt{AI-type}_{\psi}(\cdot \mid b^\texttt{AI}_t)$ is trained to maximize the expected return defined in $\mathcal{R}^\texttt{AI}$ for balancing between collaboration utility and intervention costs. For learning such policies over explicit belief representations obtained via NPF, we adopt the Proximal Policy Optimization (PPO) algorithm, which is robust to high-dimensional inputs and supports stable policy updates under approximated belief states. We provide the training details in Algorithm \ref{alg-ppo-assistance}.

\begin{algorithm}[t]
\caption{Adaptive Assistance Policy Learning via PPO and NPF}\label{alg-ppo-assistance}
\textbf{Require:} Environment simulator $\mathcal{E}^\texttt{AI}$; NPF inference model $g_{\text{npf}}$; optimal policy $\pi_*(\cdot \mid s)$; critical observation criterion; initial PPO policy parameters $\psi$; initial value function parameters $\omega$; epochs $K$; episodes per epoch $N_{\text{episodes}}$.

\begin{algorithmic}[1]
    \STATE Initialize NPF outer particles $\{\theta^{(i)}, w_0(\theta^{(i)})\}_{i=1}^{N_\theta} \sim p(\theta)$.
    \FOR{$k = 1, 2, \dots, K$}
        \STATE Initialize trajectory buffer $\mathcal{D} \leftarrow \emptyset$.
        \FOR{$e = 1, 2, \dots, N_{\text{episodes}}$}
            \STATE Sample initial state $s_0 \sim p(s_0)$, and user cognitive state $\tilde{h}_0 \sim p(\tilde{h}_0)$.
            \STATE Initialize NPF inner particles $\{\tilde{h}_0^{(i,j)}, w_0(\tilde{h}_0^{(i,j)})\}_{j=1}^{N_{\tilde{h}}} \sim p(\tilde{h}_0)$ for each $\theta^{(i)}$.
            \STATE Initialize critical observation set $O_{\text{critical}} \leftarrow \emptyset$.
            
            \FOR{$t = 0, 1, \dots, T-1$}
                \STATE Receive observation $o_t \sim \mathcal{O}(\cdot \mid s_t)$.
                \IF{$o_t$ meets critical criterion}
                    \STATE $O_{\text{critical}} \leftarrow O_{\text{critical}} \cup \{(t, o_t)\}$.
                \ENDIF
                
                \STATE Let AI observation be $o_t^\texttt{AI} \leftarrow (o_t, a_{t-1}^\texttt{H})$.
                \STATE $b_t^\texttt{AI} \leftarrow g_{\text{npf}}(b_{t-1}^\texttt{AI}, (s_t, a_{t-1}^\texttt{AI}, o_t^\texttt{AI}))$.
                
                \STATE Sample type $a_t^\texttt{AI-type} \sim \pi_{\psi}^\texttt{AI-type}(\cdot \mid b_t^\texttt{AI})$.
                \IF{$a_t^\texttt{AI-type} = \texttt{ActionHint}$}
                    \STATE $a_t^\texttt{AI} \leftarrow \arg\max_a \pi_*(\cdot \mid s_t)$.
                \ELSIF{$a_t^\texttt{AI-type} = \texttt{MemoryHint}$}
                    \STATE Sample $a_t^\texttt{AI}$ from $O_{\text{critical}}$.
                \ENDIF

                \STATE User updates cognitive state: $\tilde{h}_t \sim p(\tilde{h}_t \mid \tilde{h}_{t-1}, o_t, a_t^\texttt{AI}, \theta)$.
                \STATE User computes belief: $b_t^\texttt{H} \leftarrow p(s_t \mid \tilde{h}_t, \mathbf{a}^\texttt{H}_{:t-1})$.
                \STATE User samples action: $a_t^\texttt{H} \sim p(\cdot \mid b_t^\texttt{H}, a_t^\texttt{AI}; \theta)$.
                \STATE Environment transitions: $s_{t+1} \sim \mathcal{T}(\cdot \mid s_t, a_t^\texttt{H})$.

                \STATE Compute reward $r_t^\texttt{AI} \leftarrow r_t(s_t, a_t^\texttt{H}) - c(a_t^\texttt{AI})$.
                \STATE Store transition $(b_t^\texttt{AI}, a_t^\texttt{AI-type}, r_t^\texttt{AI}, \dots)$ in buffer $\mathcal{D}$.
            \ENDFOR
        \ENDFOR
        \STATE Update policy and value networks $(\psi, \omega)$ using PPO with data from $\mathcal{D}$.
    \ENDFOR
    \STATE \textbf{return} Trained policy $\pi_{\psi}^\texttt{AI-type}$.
\end{algorithmic}
\end{algorithm}

\end{document}